\documentclass[10pt,twocolumn,letterpaper]{article}

\usepackage[pagenumbers]{iccv}

\usepackage{pifont}
\newcommand{\cmark}{\ding{51}}
\newcommand{\xmark}{\ding{54}}

\usepackage[pagebackref,breaklinks,colorlinks]{hyperref}
\usepackage{tabularray}

\def\confName{ArXiv}
\def\confYear{2025}

\title{Practical Manipulation Model for Robust Deepfake Detection}

\author{Benedikt Hopf \hspace{3cm} Radu Timofte\\
Computer Vision Lab, CAIDAS, 
University of Würzburg, Germany\\
{\tt\small \{firstname\}.\{lastname\}@uni-wuerzburg.de}
}

\begin{document}
	\maketitle
	\begin{abstract}

Modern deepfake detection models have achieved strong performance even on the challenging cross-dataset task. However, detection performance under non-ideal conditions remains very unstable, limiting success on some benchmark datasets and making it easy to circumvent detection. Inspired by the move to a more real-world degradation model in the area of image super-resolution, we have developed a Practical Manipulation Model (PMM) that covers a larger set of possible forgeries. We extend the space of pseudo-fakes by using Poisson blending, more diverse masks, generator artifacts, and distractors. Additionally, we improve the detectors' generality and robustness by adding strong degradations to the training images. We demonstrate that these changes not only significantly enhance the model's robustness to common image degradations but also improve performance on standard benchmark datasets. Specifically, we show clear increases of $3.51\%$ and $6.21\%$ AUC on the DFDC and DFDCP datasets, respectively, over the s-o-t-a LAA backbone. Furthermore, we highlight the lack of robustness in previous detectors and our improvements in this regard. Code can be found at \url{https://github.com/BenediktHopf/PMM}.
\end{abstract}  
	\begin{figure}
    \centering
    \begin{subfigure}{0.8\linewidth}
        \centering
        \includegraphics[width=\linewidth]{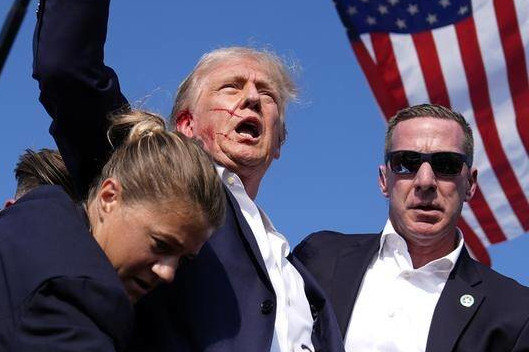}
        \caption{\textcolor{green}{Real} image~\protect\cite{moreechampion2024trump_asassination}: LAA~\protect\cite{nguyen2024laa} predicts \textcolor{green}{real}, ours \textcolor{green}{real}.}
        \label{fig:teaser_real}
    \end{subfigure}
    \begin{subfigure}{0.8\linewidth}
        \centering
        \includegraphics[width=\linewidth]{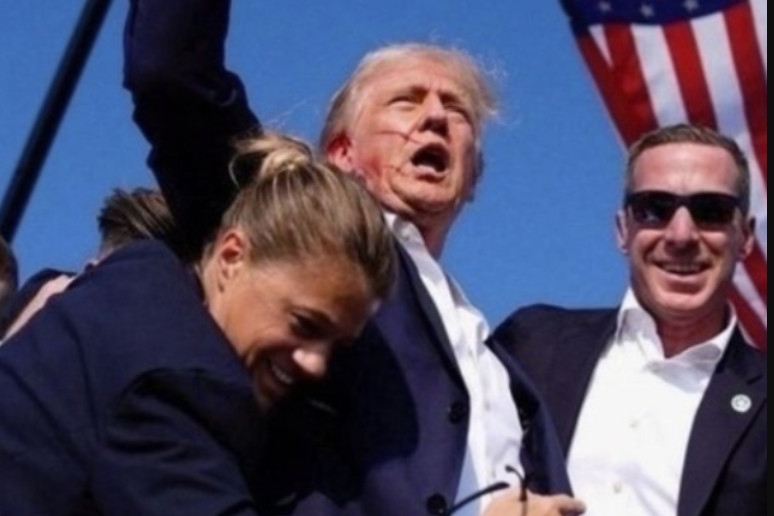}
        \caption{\textcolor{red}{Fake} image~\protect\cite{dfrlab2024trump_asassination}: LAA~\protect\cite{nguyen2024laa} predicts \textcolor{green}{real}, ours \textcolor{red}{fake}.}
        \label{fig:teaser_fake}
    \end{subfigure}
    \caption{
    \textbf{Our method performs robustly on `in the wild' images: }
    A \textcolor{green}{real} image~\protect\cite{moreechampion2024trump_asassination} from the assassination attempt on Donald Trump and a corresponding altered / \textcolor{red}{fake} image~\protect\cite{dfrlab2024trump_asassination} with faked smiles. \textit{Note that the fake image has not been altered by us but is available online like this.} Our method works well whereas the s-o-t-a LAA~\protect\cite{nguyen2024laa} method fails to detect the deepfake. 
    }

    \label{fig:teaser}
\end{figure}

\section{Introduction}
\label{sec:intro}

The widespread availability of high-quality generative methods for fake image creation has led to numerous \textit{deepfakes} on the Internet and a significant social impact~\cite{ahmed2021inadvertently,hancock2021social}. In response, deepfake detection has gained significant attention to address societal issues like fraud and fake news. Early detectors~\cite{Afchar2018MesoNetAC, Jung2020Blinking, roessler2019faceforensicspp, Sabir2019RecurrentCS} perform well when test images match the training distribution, but this is unrealistic in real-world scenarios. Consequently, cross-dataset evaluation~\cite{Cozzolino2018ForensicTransferWD, Li2019FaceXRay, Shiohara2022SBI, nguyen2024laa} has become the standard for assessing modern detectors, measuring real-world generalization across forgery techniques and source data. However, this protocol does not account for the vulnerability of detectors to low-quality images as shown in our study (see \cref{sec:experiments}).

In line with prior research~\cite{ahmed2021inadvertently,hancock2021social}, we argue that there are two types of deepfakes: (i) non-deceiving deepfakes, often used on movies and typically labeled as fake, and (ii) malicious deepfakes, created to deceive viewers. While both aim for high-quality fakes, only type (ii) seeks to avoid detection. This can be achieved by using low-quality images (\eg, noisy and low resolution, see \cref{sec:robustness}) and providing a plausible explanation for the poor quality (\eg, shot on an old mobile phone in the dark). In such cases, it is crucial that deepfake detectors can recognize these images. An example of a real-world low-quality fake image, which avoids detection by being slightly blurry, is shown in \cref{fig:teaser_fake}. This image serves as exemplary proof that the described effect is getting exploited. Since it is non-trivial to collect these types of images in large quantities from the web (as they are not labeled as deepfakes), we will manually create such conditions in quantitative experiments.

Since type (ii) fakes are, of course, much more dangerous, we argue that it is of paramount importance to also address robustness in the analysis of state-of-the-art models. This is backed by the fact that the two benchmark datasets DFDC~\cite{Dolhansky2020DFDC} and DFDCP~\cite{Dolhansky2019DFDCP}, which present more difficult conditions, show strong improvements as a result of our method (see \cref{sec:sota}). Even though image quality will most likely continue to increase with time, low-quality media will still be produced, especially by cheaper devices and under challenging conditions. We therefore expect the problem to stay relevant in the future.

A similar move from a limited academic setting to a more real-world setting has been performed in the area of single image super-resolution~\cite{Zhang2021BSRGAN}. Inspired by this, we propose to diversify the data that the detector receives during training. We include very low-quality images in the training and also extend the popular framework of self-blended images (SBI)~\cite{Shiohara2022SBI} with more diverse masks, Poisson blending, distractors, and generator artifacts.

These changes allow much more stable performance toward image degradations as well as a significant improvement on the DFDC and DFDCP~\cite{Dolhansky2020DFDC, Dolhansky2019DFDCP} benchmark datasets. We achieve this without changing model architecture, just by using a more diverse set of pseudo-fakes. We base the experiments in this paper on LAA~\cite{nguyen2024laa}, which was proposed to detect high-quality deepfakes. We extend its capabilities by allowing it to also detect deepfakes in low-quality images and, therefore, bring it a step closer to actually being usable for detecting deepfakes in the real world.

The only other works we are aware of that explore image quality for deepfake detection in detail are~\cite{lu2022ddrc, Lu2023AssessmentFF}. However, they use much less general degradations, and do not provide strong benchmark performance. Furthermore, \cite{Yan2023DeepfakeBenchAC, nguyen2024laa, Cao2022RECCE, Yan2023LSDA} analyze robustness to image degradations but do not propose any changes based on their analysis. 

While Poisson blending and new mask shapes are additions to the known process of blending pseudo-fakes, our purposeful introduction of a variable amount of generator artifacts is, to the best of our knowledge, unprecedented.

\vspace{2mm}
\noindent\textbf{Our main contributions} can be summarized as follows: 
\begin{enumerate}
    \item[1)] Extensions to the popular framework of SBI~\cite{Shiohara2022SBI}, for more general pseudo-fakes.
    \item[2)] A powerful new and purpose-built degradation model to greatly improve the robustness to many kinds of image degradations, going far beyond the capabilities of previous degradation models.
    \item[3)] Specifically modelling generator artifacts in pseudo-fakes, which are usually free of such effects.
    \item[4)] Extensive experiments to show how our proposed Practical Manipulation Model (PMM) adds robustness to the s-o-t-a deepfake detectors and improves performance on low-quality images and standard benchmarks.
\end{enumerate}

\noindent To the best of our knowledge, our paper is the first to analyze the failure cases of modern deepfake detection methods in terms of image degradations and to provide a solution that not only significantly improves performance in this regard but even on standard benchmark datasets.
Thus, we consider our work to advance the understanding of the deepfake field. Our model goes beyond a trivial combination of prior methods by creating a combination that achieves improvements in standard benchmarks. Direct use of the degradation model from~\cite{Zhang2021BSRGAN} designed for blind restoration, for example, leads to poor results ($80.9\%$ vs. $93.2\%$ AUC on DFDCP). Furthermore, we are the first to model generator artifacts in pseudo-fake generation, which significantly improves performance, despite the training generators being different from the ones used in the test set.

	\section{Related Work}
\label{sec:relatedwork}

\noindent\textbf{Deepfake detection.} To combat the spread of unrecognized deepfakes, several works have focused on detection. Methods range from simple binary classification~\cite{roessler2019faceforensicspp} over frequency-based methods~\cite{Liu2021SPSL, Luo2021GeneralizingFFHighFrequencyDetails, Hasanaath2024FSBI, Masi2020TwobranchRN, Qian2020F3Net} to spatial methods~\cite{Yan2023UCFUC, Yan2023LSDA, Cao2022RECCE, He2021BeyondTS, Guo2023ControllableGF, Zhuang2022UIAViTUI}. Such methods rely on specific properties of deepfakes to aid detection. The CNNs that are usually used to generate deepfakes, for example, leave artifacts in the frequency domain~\cite{Wang2019CNNGeneratedIA}. Furthermore, \cite{Cao2022RECCE}, for example, uses reconstructions from autoencoders.

Other methods~\cite{Li2019FaceXRay, Zhao2020PCL, Shiohara2022SBI, nguyen2024laa, Larue2022SeeABLESD} build pseudo-fakes -- faceswaps without the need for a generator, by either choosing well-matching videos or even modified copies of the same video -- to avoid overfitting to generator artifacts. This, however, has the downside of not being able to use generator artifacts at test time, despite their prominent use in frequency-based detectors. We therefore provide an extension to this class of methods, which achieves the best of both worlds. We use the network architecture from~\cite{nguyen2024laa} -- a current s-o-t-a method with public code -- as the basis of our method, because we do not change the model architecture.

\vspace{2mm}
\noindent\textbf{Degradation modeling.}
While identifying various types of deepfakes is crucial for practical applications, other challenges may arise in real-world scenarios. One notable problem is image degradation. The effects of image quality on deepfake detectors have been studied by~\cite{lu2022ddrc, Lu2023AssessmentFF}. They provide a simple degradation model and test the effects of training on it. However, their degradation model is less general than ours, they do not consider blending, generator artifacts, distractors, and masks. Furthermore, they base their work on and compare it to outdated methods and achieve relatively weak benchmark performances. 

Some methods~\cite{nguyen2024laa, Cao2022RECCE, Yan2023LSDA, Yan2023DeepfakeBenchAC} test their models against image degradations, but this is always just an analysis with no or very limited consequences and recommendations.

In the field of image super-resolution, degradation modeling is a crucial topic. Our method is inspired by \cite{Zhang2021BSRGAN}, which achieves superior results in the wild by the use of a more complex degradation model. While the application of their method to deepfake detection does not result in good performance (see \cref{sec:intro}), due to being built for a different task, we can still include some of their ideas.
	\section{Proposed Method}
\label{sec:method}

\begin{figure}
    \centering
    \newcommand{\pmmExampleScale}{0.12}
    \begin{tabular}{|cc|cc|}
    \hline
        \multicolumn{2}{|c|}{LAA~\cite{nguyen2024laa}} & \multicolumn{2}{c|}{ours} \\ \hline
        real & fake & real & fake \\ 
         \includegraphics[scale=\pmmExampleScale]{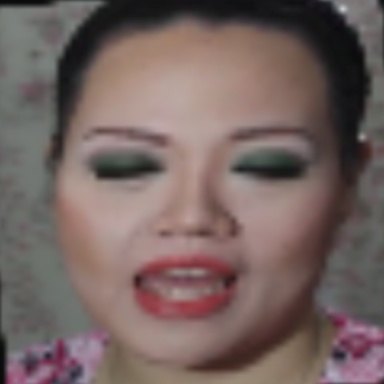} &
         \includegraphics[scale=\pmmExampleScale]{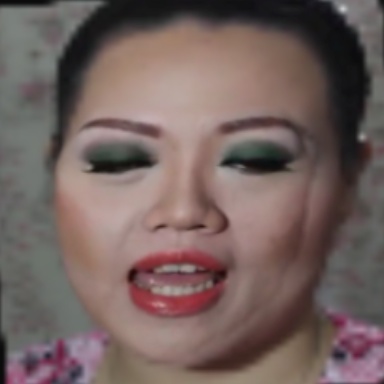} &
         \includegraphics[scale=\pmmExampleScale]{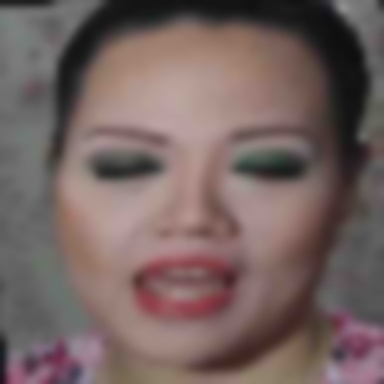} & 
         \includegraphics[scale=\pmmExampleScale]{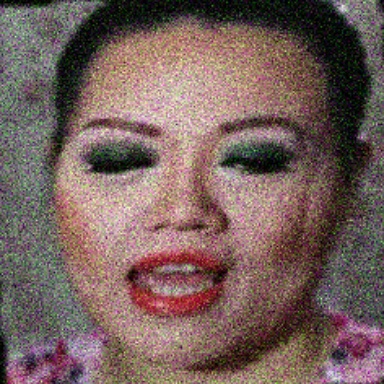} \\
         \includegraphics[scale=\pmmExampleScale]{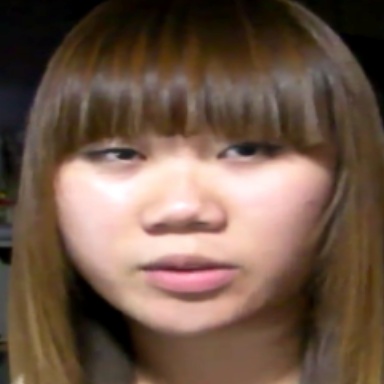} &
         \includegraphics[scale=\pmmExampleScale]{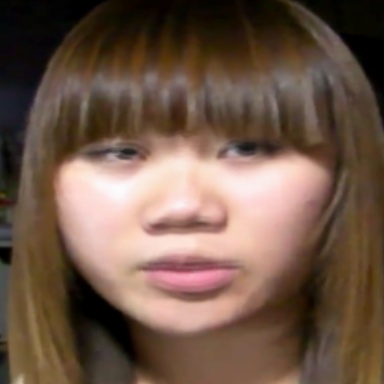} &
         \includegraphics[scale=\pmmExampleScale]{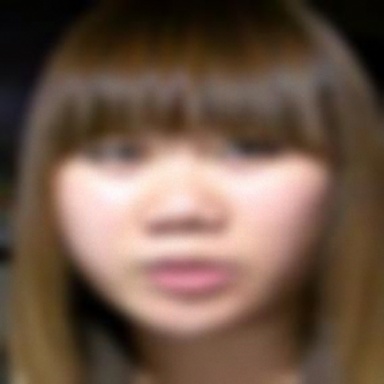} & 
         \includegraphics[scale=\pmmExampleScale]{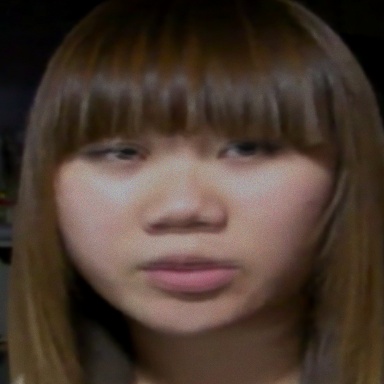} \\
         \hline
          
    \end{tabular}
    \caption{\textbf{Examples of our training images} compared to the ones from LAA~\protect\cite{nguyen2024laa}. Our images are much harder to detect due to their strong augmentations. The original images are taken from the FaceForensics++ dataset~\protect\cite{roessler2019faceforensicspp}.}
    \label{tab:pmm_examples}
\end{figure}

We propose a Practical Manipulation Model (PMM) to improve the robustness of a deepfake detector to the aforementioned weaknesses. At the same time, we demonstrate that our proposed data generation strategy brings a significant improvement on the DFDC and DFDCP~\cite{Dolhansky2020DFDC, Dolhansky2019DFDCP} datasets, surpassing the previous s-o-t-a by a significant margin. Example images of our method can be seen in \cref{tab:pmm_examples}.

We achieve this by only changing the data preparation, without any changes to the model architecture. Our approach is, therefore, applicable to any deepfake detector. We base the evaluations in this paper on the Localized Artifact Attention (LAA) model~\cite{nguyen2024laa} for its state-of-the-art performance, particularly on high-quality deepfakes. Additionally, we also test EfficientNet-b4~\cite{Tan2019EfficientNetRM} and Xception~\cite{Chollet2016XceptionDL}, compared to their SBI~\cite{Shiohara2022SBI} versions.

\begin{figure*}
    \centering
    \includegraphics[width=\linewidth]{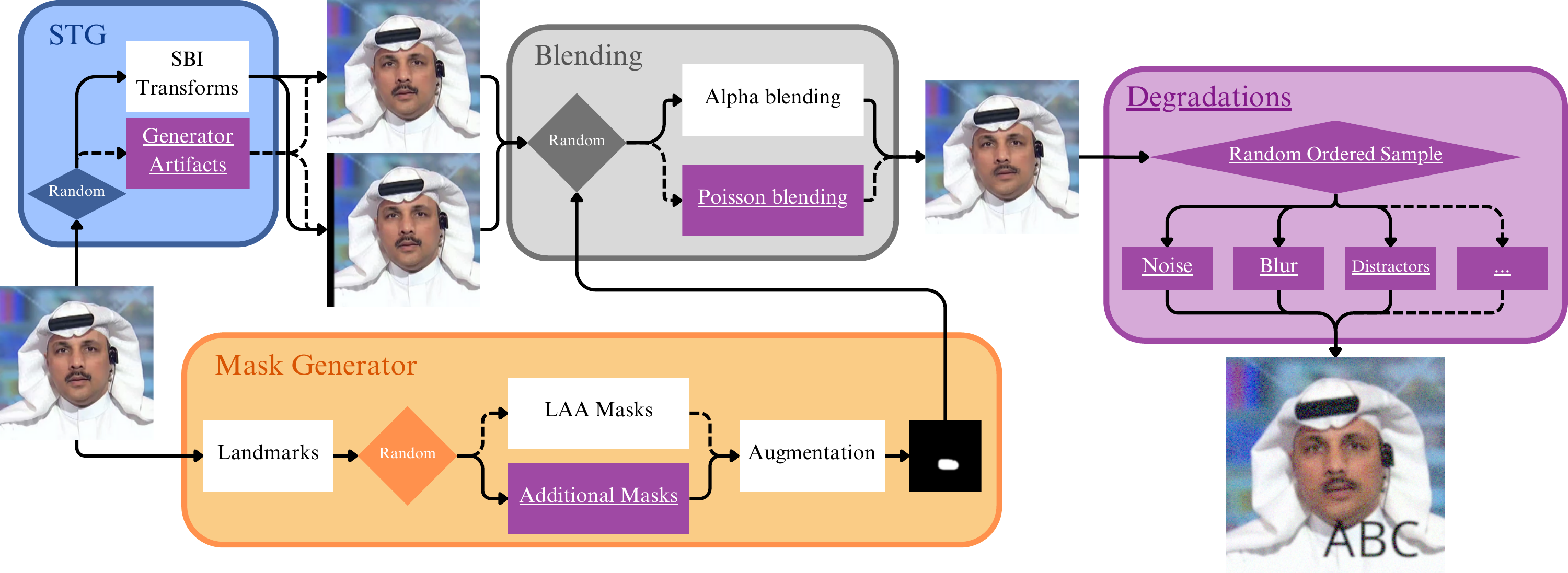}
    \caption{\textbf{Overview of our proposed Practical Manipulation Model (PMM).} Our model extends the self-blending from~\protect\cite{Shiohara2022SBI} by considering generator artifacts, Poisson blending, additional masks, and degradations, all denoted as purple elements and underlined. Random choices can take any path. Dashed lines indicate paths not taken in the example.}
    \label{fig:scheme} 
\end{figure*}

As shown in \cref{fig:scheme}, we make contributions to the entire SBI pipeline (compare to Fig.~3 in \cite{Shiohara2022SBI}). For the Source-Target Generator (STG), we add the option of using generator artifacts to distinguish the images. The Mask Generator is improved by additional masks, and the blending step can also perform Poisson blending. Finally, we apply our degradation model to the finished pseudo-fake.

We perform all preprocessing steps asynchronously on the CPU, as is commonly done for preprocessing. Therefore, provided enough CPU resources are available, no slowdown can be observed. 
Since training is usually GPU-bound, the computational overhead is negligible, which aligns with our measurements.

\subsection{Degradation model}
\label{sec:degradation_model}

Since deepfake detectors have to pick up on, in many cases, almost imperceptible artifacts, they tend to be vulnerable to image degradations. While color-based changes, like contrast or saturation shifts, usually do not hinder detectors much, as they leave the structure of the image the same, degradations like noise, compression, or blur can significantly reduce the performance of s-o-t-a detectors~\cite{Lu2023AssessmentFF, nguyen2024laa, lu2022ddrc}.
Usually, this topic is considered from the point of image manipulations that are \textit{benignly} performed on an image for some unrelated reason. However, this property of deepfake detectors could also be used \textit{maliciously} to circumvent detection. Adding a little bit of noise might have a similar effect to an adversarial attack~\cite{Szegedy2013AdversarialAttacks} but requires no access to the model.
Therefore, it is of paramount importance to strengthen the robustness of deepfake detectors to as broad a set of perturbations as possible. 

Inspired by the \textit{practical degradation model} from Zhang~\etal~\cite{Zhang2021BSRGAN}, we propose to build a degradation model specifically designed for deepfake detection. We use some of their individual degradations but include additional ones and remove others. One main contribution of their work is the idea of applying degradations in random order. This significantly increases the space of possible results, providing more variety during training and better generalization. 

By carefully choosing degradations, such that they are hard, but still allow for detection, we can use stronger degradations than previous methods, leading to stronger benefits. One example of this is the specific use of strong noise ($\sigma > 40$). This still allows for detection, because many pixels still have close-to-original values. But it also introduces very strong high-frequency artifacts, which the model can learn not to use as a (bad) detection criterion.

Since we are not performing restoration but deepfake detection, we apply all the degradations only with some probability $p$. This gives the model the chance to learn from high-quality, non-degraded images as well as low-quality ones. Otherwise, all images would always be degraded, which is realistic for a restoration model but not for a deepfake detector.  All degradations are individually sampled, such that any combination of degradations can appear during training, similar to~\cite{lu2022ddrc, Lu2023AssessmentFF}. Our list of degradations is as follows:

\vspace{2mm}\noindent\textbf{Smoothing.}
Smoothing is performed by convolution with a blur kernel. This kernel can be created in one of three ways (uniformly chosen): 
\begin{enumerate}
    \item Anisotropic Gaussian blur. Both standard deviations are randomly sampled from $Uniform(0, 30\cdot s)$, and the angle from $\mathit{Uniform}(0, \pi)$.
    \item The \texttt{f\_special} method from the implementation of \cite{Zhang2021BSRGAN}, with standard deviation chosen from $\mathit{Uniform}(0, 30\cdot s)$.
    \item A uniform kernel $k_{ij} = \frac{1}{w^2}$ for some square size $w \sim \mathit{Uniform}(3, 30\cdot s)$ and $i, j \in \{1, ..., w\}$.
\end{enumerate}

\vspace{2mm}\noindent\textbf{Resize.}
For resizing, there are also three options: \begin{enumerate}
    \item Upsampling with a scale factor $\mathit{Uniform}(1, 2)$.
    \item Downsampling with a scale factor $\mathit{Uniform}\left(\frac{0.25}{s}, 1\right)$.
    \item Resampling with a scale factor of $1$.
\end{enumerate}
Exactly one of the three options is applied, with probabilities $0.2, 0.7$, and $0.1$ respectively. The resampling method is uniformly chosen from linear, cubic, and area.

\vspace{2mm}\noindent\textbf{Gaussian noise.}
There are three options: 
\begin{enumerate}
    \item Uncorrelated Gaussian noise for all channels with standard deviation chosen from $\mathit{Uniform}(l_1\cdot s, l_2 \cdot s)$.
    \item Grayscale Gaussian noise (\ie the same for all three color channels) with standard deviation chosen from $\mathit{Uniform}(l_1\cdot s, l_2 \cdot s)$.
    \item Channel-wise correlated noise, where the covariance matrix for the three image channels is not diagonal.
\end{enumerate}
The three options are applied with probabilities $0.4, 0.4, 0.2$, respectively. Since we particularly want to focus on strong noise -- a large problem for deepfake detection -- Gaussian noise is applied either for $l_1=2, l_2=100$ or $l_1=80, l_2=100$. To increase variety even more, the options are not exclusively chosen, but can be applied at the same time. Rather than applying Gaussian noise with probability $p$, both cases are independently applied with probability $\frac{p}{2}$.

\vspace{2mm}\noindent\textbf{Non-Gaussian noise.}
For non-Gaussian noise, we use either speckle or Poisson noise. Which one is randomly selected with probability $\frac{1}{2}$.
Speckle noise is calculated exactly as Gaussian noise, but multiplied with the image before adding it to the image. The noise levels are chosen to be $l_1=2, l_2 = 25$.
Poisson noise is applied unchanged from the version in the code of~\cite{Zhang2021BSRGAN}.

\vspace{2mm}\noindent\textbf{JPEG compression.} We add JPEG noise for compression with a quality factor chosen from $\mathit{Uniform}(10, 95)$.

\vspace{2mm}\noindent\textbf{Enhance.} As in~\cite{lu2022ddrc, Lu2023AssessmentFF}, we enhance the image by changing either Brightness or Contrast. We use the \texttt{ImageEnhance}-module provided by \texttt{Pillow}~\cite{clark2015pillow}, with enhancement factor from $\mathit{Uniform}(0.5, 1.5)$.

\vspace{2mm}\noindent\textbf{Distractors.} We add either random text or a random image from the same dataset to the training image in order to simulate situations in the real world, where the video might have an ad or some other kind of overlay on it. Similar distractors also appear in the DFDC~\cite{Dolhansky2020DFDC} dataset.
Images are randomly placed in the image, resized to $x \sim \mathit{Uniform}(20, 100)$ pixels in the $x$ direction and $y \sim \mathit{Uniform}(x\cdot 0.8, x\cdot 1.2)$ in the $y$ direction.
Text is overlayed at a random position using the \texttt{putText} method from~\cite{opencv_library}, where all the parameters are randomly populated for maximum variability.
The distractors are applied with a separate probability $p_d$. Multiple distractors can be added to a single image. For details, we refer to the supplementary material.

\vspace{2mm}\noindent\textbf{Application of Degradations.} Degradations are applied in random order and, unless otherwise mentioned, with a probability of $p$ each. If the image was resized as part of the degradation process, it is resized back to the original resolution after all degradations are done. Random order and random application significantly increase the space of possible results, which provides more variety during training.

\subsection{Blending}

The blending step is very important, as it is common to all non-entire-face-synthesis deepfakes. Methods like~\cite{Li2019FaceXRay, Shiohara2022SBI} rely entirely on effects at the blending boundary. We add variety to the blending itself and introduce an entirely new method for creating the to-be-blended base data, in the form of generator artifacts.

\vspace{2mm}\noindent\textbf{Blending method.}
Adding to the commonly used alpha blending, given by $I = M\cdot S + (1-M) \cdot T$ for some source image $S$, target image $T$, and mask $M$, we add the option of Poisson blending~\cite{Prez2003PoissonIE}. This method is also very capable of creating realistic deepfakes, as can be seen by its usage in \cite{Dolhansky2020DFDC}. We randomly choose Poisson blending over alpha blending with a probability of $p_p$.

\vspace{2mm}\noindent\textbf{Blending mask.}
LAA~\cite{nguyen2024laa} already uses different kinds of masks for blending, however, they all focus on the entire face. To add more variety, we add another type of mask that is based on semantic parts of the image (different from the grid parts in \cite{Larue2022SeeABLESD}). One part of the face is randomly uniformly chosen from: right jaw, left jaw, right cheek, left cheek, nose ridge, nose, right eye, or left eye. Only this specific part is replaced in the pseudo-fake, rather than the entire face. We keep the masking strategies from LAA and uniformly sample one of them or our semantic one.

\vspace{2mm}\noindent\textbf{Generator Artifacts.}
SBI~\cite{Shiohara2022SBI} usually creates a detectable blending boundary by applying some transform (color or geometric) to generate either the source or target image. We argue that generator artifacts should also be modeled for increased variety, and introduce a new way to do that, without losing the advantages of pseudo-fakes. Instead of the SBI transforms, with probability $p_g$, we introduce artifacts generated by either stable diffusion~\cite{rombach2021stablediffusion} or the GAN from~\cite{bobkov2024ganediting}. The images are created by the \texttt{img2img} method of stable diffusion with a denoising strength of $0.01$ in order to barely have any effect on the content of the image but to introduce the artifacts of stable diffusion. The images from~\cite{bobkov2024ganediting} are inverted for editing, but the editing strength is set to $0$ for the same reason.

For every image in the training dataset, we generate one image by each of these methods as a data preparation step before training. This is \textbf{not} intended to be a new dataset but rather a cache of generator artifacts. Ideally, we would generate these images at training time and different for each epoch (just like the SBI transforms), but loading two large image generators in addition to the actual training is computationally too prohibitive. We, therefore, resort to caching.

When generating the self-blended image, we use each generator with probability $p_g$ and otherwise use the standard SBI implementation in the code of LAA~\cite{nguyen2024laa}, directly controlling the extent to which the model is shown generator artifacts. 
	\section{Experiments}
\label{sec:experiments}

\begin{table}
    \centering
    \resizebox{\linewidth}{!}
    {
    \begin{tabular}{|c|ccc|c|}\hline
         Method & 
         CDFv2 & DFDC & DFDCP & Average  \\ \hline \hline
         Xception~\cite{roessler2019faceforensicspp, nguyen2024laa} &  0.6118 & - & 0.6990 & \textcolor{gray}{0.6554} \\
         EfficientNet-B4~\cite{Tan2019EfficientNetRM, Yan2023DeepfakeBenchAC} & 0.7487 & 0.6955 & 0.7283 & 0.7242 \\
         FaceXRay+BI~\cite{Li2019FaceXRay} & 
         0.7950 & 0.6550 & 0.8092 & 0.7531\\
         RECCE~\cite{Cao2022RECCE} & 
         0.6871 & 0.6906 & - & \textcolor{gray}{0.6889}\\ 
         SPSL~\cite{Liu2021SPSL} & 
         0.7688 & 0.7040 & 0.7408 & 0.7379\\
         
         UIA-ViT~\cite{Zhuang2022UIAViTUI} & 
         0.8241 & - & 0.7580 & \textcolor{gray}{0.8203}\\
         SeeABLE~\cite{Larue2022SeeABLESD} & 0.8730 & 0.7590 & 0.8630 & 0.8317 \\
         UCF~\cite{Yan2023UCFUC,Yan2023DeepfakeBenchAC} &
         0.7527 & 0.7191 & 0.7594 & 0.7437\\
         Controllable GS~\cite{Guo2023ControllableGF} & 
         0.8497 & - & 0.8165 & \textcolor{gray}{0.8331}\\
         LSDA$^\text{frame-level}$~\cite{Yan2023LSDA} & 
         0.830 & 0.736 & 0.815 & 0.7937\\
         \hline
         SBI (Xception)~\cite{Shiohara2022SBI} &
         0.8292* & 0.7050* & 0.8193* & 0.7845\\
         +PMM (ours) & 
         0.8015 & \underline{0.7692} & 0.8465 & 0.8056\\ 
         
         \hline
         SBI (EfficientNet)~\cite{Shiohara2022SBI} &
         \underline{0.9318} & 0.7242 & 0.8615 & 0.8392\\
         +PMM (ours) & 
         0.8927 & \textbf{0.7878} & \underline{0.8794} & \underline{0.8533}\\
          
         \hline
         LAA~\cite{nguyen2024laa} & 
         \textbf{0.9540} & 0.7170* & 0.8694 & 0.8468\\
         +PMM (ours) & 
         0.9153 & 0.7521 & \textbf{0.9315} & \textbf{0.8663}\\ \hline
    \end{tabular}
    }
    \caption{\textbf{Cross-dataset evaluation and comparison to the state-of-the-art in terms of AUC ($\uparrow$).} Best values are highlighted in \textbf{bold}, second best values are \underline{underlined}. Results marked with a * are from our own testing of the official source code. \textcolor{gray}{Gray} values indicate averages based on an incomplete set of benchmarks.}
    \label{tab:sota}
\end{table}

\begin{table*}
	\centering
	\begin{tabular}{|c||c|c|c|c|c||c|c|} \hline
		& \multicolumn{5}{|c||}{training setting} & \multicolumn{2}{c|}{AUC($\uparrow$)}\\ \hline
		Variant & Degradations & Add. Masks & Poisson Blending & Distractors & Gen. Art. & Plain & Low-Quality \\ \hline \hline
		1 (LAA)&  &  &  &  &  & 0.8413 & 0.7648 \\ 
		2 & \cmark &  & &  &  & 0.9196 & 0.9143 \\ 
		3 & \cmark & \cmark &  &  &  & 0.9198 & 0.9111 \\ 
		4 & \cmark & \cmark & \cmark &  & & \underline{0.9201} & \textbf{0.9229} \\ 
		5 & \cmark & \cmark & \cmark &  \cmark & & 0.9035 & 0.9097 \\ 
		6 (LAA+PMM) & \cmark & \cmark & \cmark & \cmark & \cmark & \textbf{0.9315} & \underline{0.9194} \\ 		
		\hline
	\end{tabular}
	
	\caption{\textbf{PMM ablative results on DFDCP~\protect\cite{Dolhansky2019DFDCP} (cross-dataset).} Variant 6 is our default PMM method, Variant 1 corresponds to the LAA backbone. Balancing between Plain and Low-Quality settings, we choose variant 6 as our final model. It achieves the best average AUC of $0.9258$, compared to $0.9212$ of variant 4.}
	\label{tab:ablation}
\end{table*}

In this section, we describe the settings under which we performed our experiments, compare our model to the state-of-the-art, and evaluate the effects of different backbones. Furthermore, we perform an ablation study of our design choices and analyze the increase in robustness obtained through our method. Additional experiments (including GradCAM~\cite{Selvaraju2016GradCAMVE} to show the improved focus of our models) are provided in the supplementary.

\subsection{Experimental settings}

\vspace{2mm}\noindent\textbf{Datasets.} We follow the usual evaluation protocol of cross-dataset testing~\cite{nguyen2024laa, Shiohara2022SBI, Yan2023LSDA, Yan2023UCFUC, Cao2022RECCE, Liu2021SPSL, Yan2023DeepfakeBenchAC}. We train on the FaceForensics++ dataset~\cite{roessler2019faceforensicspp}. The dataset contains a total of 5000 videos across its training, validation, and testing splits. There are 1000 pristine videos and another 1000 videos from each of the four used manipulation methods: Deepfakes~\cite{deepfakes}, FaceSwap~\cite{faceswap}, Face2Face~\cite{Thies2016Face2Face}, and Neural Textures~\cite{Thies2019NeuralTextures}. Given that our method is based on LAA~\cite{nguyen2024laa}, which is based on SBI~\cite{Shiohara2022SBI}, we only use real images during training and blend our own pseudo-fakes according to the previously described approach (see \cref{fig:scheme}).

For the cross-dataset scenario, we mostly use the DFDCP~\cite{Dolhansky2019DFDCP} dataset. We consider this the most useful for our experiments as it includes different image qualities, making it a more realistic approximation to an actual real-world scenario than the higher-quality datasets. DFDC~\cite{Dolhansky2020DFDC} is a larger dataset and arguably even better suited for this task, so we reserve this as an unseen test-set, which we do not use during ablation, but only evaluate the final model on. We also report values on Celeb-DF-v2 (CDFv2)~\cite{Li2019CelebDFv2} because it is one of the most common benchmark datasets, even though it is less relevant to our hypothesis. Furthermore, we perform one test on a subset of DF40~\cite{yan2024df40}

\vspace{2mm}\noindent\textbf{Methods.}
We test our proposed method against twelve deepfake detection models, ranging from baselines to state-of-the-art methods: Xception~\cite{Chollet2016XceptionDL, roessler2019faceforensicspp}, EfficientNet-B4~\cite{Tan2019EfficientNetRM}, FaceXRay~\cite{Li2019FaceXRay}, RECCE~\cite{Cao2022RECCE}, SPSL~\cite{Liu2021SPSL}, UIA-ViT~\cite{Zhuang2022UIAViTUI}, SeeABLE \cite{Larue2022SeeABLESD}, UCF~\cite{Yan2023UCFUC}, Controllable Guide-Space~\cite{Guo2023ControllableGF}, LSDA~\cite{Yan2023LSDA}, SBI~\cite{Shiohara2022SBI} and LAA~\cite{nguyen2024laa}. Adhering to common practice~\cite{nguyen2024laa, Shiohara2022SBI, Yan2023LSDA, Yan2023UCFUC, Cao2022RECCE, Liu2021SPSL, Yan2023DeepfakeBenchAC}, we test the methods under the cross-dataset protocol and cite their official numbers where available. Otherwise, we cite the numbers from some other paper, which has numbers for the specific method.

\vspace{2mm}\noindent\textbf{Measures.} We mainly report the area under the receiver operating characteristic curve (AUC) as it is commonly done in related work~\cite{Shiohara2022SBI, nguyen2024laa, Cao2022RECCE, Liu2021SPSL, Yan2023DeepfakeBenchAC, Yan2023UCFUC, Yan2023LSDA}. AUC has the advantage that it is a parameter-free metric and does not depend on the detection threshold, like, \eg, accuracy would. 

\vspace{2mm}\noindent\textbf{Implementation details.} We stick closely to the settings of LAA with minor changes (see supplementary for details).

\subsection{Comparison to the state-of-the-art}
\label{sec:sota}

We compare our method to the state-of-the-art using the standard cross-dataset evaluation protocol as in~\cite{nguyen2024laa}. We train our model on FaceForensics++~\cite{roessler2019faceforensicspp} and test on other datasets: CDFv2~\cite{Li2019CelebDFv2}, DFDC~\cite{Dolhansky2020DFDC}, and DFDCP~\cite{Dolhansky2019DFDCP}.

\Cref{tab:sota} shows the methods' results as reported by the original authors or in cited works. 
The table shows that while we lose some performance on the high-image-quality dataset CDFv2, we have significant gains on the lower-quality datasets DFDC and DFDCP. This result was expected, as our model is intended for robustness on low-quality images, which, in turn, reduces the model's capacity for the high-quality task (see \cref{sec:limitations}). Due to the large improvement on the other two datasets, however, the average improves by $\approx2\%$ AUC.

\vspace{2mm}\noindent\textbf{Different Backbones.}
Since our Practical Manipulation Model (PMM) is applicable to any backbone architecture, we additionally validate its effectiveness on EfficientNet-b4~\cite{Tan2019EfficientNetRM} and Xception~\cite{Chollet2016XceptionDL} backbones, compared to training under the standard SBI~\cite{Shiohara2022SBI} settings. As shown in \cref{tab:sota}, there are similar effects to the ones using LAA~\cite{nguyen2024laa}. We observe a slight decrease in AUC performance on CDFv2 and significant improvements, $+6.36\%$ and $+1.79\%$ (EfficientNet), on the harder datasets DFDC and DFDCP, respectively. On DFDC, in particular, our PMM with EfficientNet achieves the best AUC of $0.7878$. The Xception backbone achieves an average improvement of $2.11\%$. This validates our hypothesis that PMM can be applied to different backbone architectures for improved robustness to degradations and underlines the versatility of our PMM approach.

\subsection{Ablation study}
\label{sec:ablation}

In this section, we evaluate the effect of the individual parts of our proposed Practical Manipulation Model (PMM) and present them in \cref{tab:ablation}. The results are reported on the DFDCP~\cite{Dolhansky2019DFDCP} dataset, while all models/variants are trained on the original images from FaceForensics++~\cite{roessler2019faceforensicspp} and generated pseudo-fakes. We report AUC according to the standard protocol in the area of deepfake detection~\cite{Shiohara2022SBI, nguyen2024laa, Cao2022RECCE, Liu2021SPSL, Yan2023DeepfakeBenchAC, Yan2023UCFUC, Yan2023LSDA}.

We start from the configuration of LAA~\cite{nguyen2024laa} (Variant 1 (LAA) in \cref{tab:ablation}) and progressively add the individual parts of our proposed method: Degradations, Additional Masks, Poisson Blending, Distractors, and Generator Artifacts (Gen.Art.). Since we want to ensure robustness, we measure performance under two different conditions: \textit{Plain} and \textit{Low-Quality}. \textit{Plain} uses the unchanged images from the DFDCP dataset. In addition, we run all the tests a second time, applying a slightly weakened version of the degradations from \cite{Zhang2021BSRGAN}, resulting in \textit{Low-Quality} images. This allows us to also consider the model's performance under real-world, low-quality conditions. Results are presented in \cref{tab:ablation}; more details are in the supplementary material.

\vspace{2mm}\noindent\textbf{Degradations.} 
Variants 1 (LAA) and 2 in \cref{tab:ablation} show the effects of using our degradation model during training. As expected, the original model (Variant 1) loses $7.65\%$ AUC on Low-Quality images. With the addition of training-time degradations, the performance increases for the Plain setting to $91.96\%$ AUC. Robustness also greatly profits, with the AUC performance drop reduced to only $0.53\%$.

\vspace{2mm}\noindent\textbf{Additional Masks.} We see a slight improvement in AUC with additional masks in Variant 3 \wrt Variant 2. To the best of our knowledge, DFDCP~\cite{Dolhansky2019DFDCP} does not use partial masks. Therefore, we do not expect a large improvement here but would rather expect increased generalization ability when tested on a dataset with such properties. 

\vspace{2mm}\noindent\textbf{Poisson Blending.} Adding Poisson blending (\ie, Variant 4) results in another slight boost to the performance on the Plain dataset. When tested with degradations, the advantage is much larger, resulting in the best AUC of $0.9229$. Here, the AUC tested on degraded DFDCP is even slightly better than without degradations. We hypothesize that the degradations hide some smaller artifacts in the image, forcing the model to focus on more generalizable signs of a fake. If this effect is stronger than the added information in the Plain images, a situation like this can arise.

\vspace{2mm}\noindent\textbf{Distractors.} In Variant 5, we add distractors to improve robustness in real-world settings. Since the DFDCP dataset does not use distractors, and neither do the degradations from \cite{Zhang2021BSRGAN}, our Plain and Low-Quality test settings cannot capture this robustness. Therefore, we do not expect to see an improvement under these test settings. However, we showcase the positive effect of this addition in \cref{sec:qualitative_demonstration}.

\vspace{2mm}\noindent\textbf{Generator artifacts.} Finally, we test the usage of generator artifacts in the full PMM (\ie, Variant 6), which provides the detector with an additional way to detect fake images, which is usually missing in blending-based methods. We see a strong improvement in both test settings. The best AUC is achieved under the Plain test setting for our complete method with an AUC of $0.9315$. Note that neither of the two generators used during training is part of the test sets. This result, therefore, shows generalization to different generators, despite the limited variety during training.

\subsection{Robustness analysis}
\label{sec:robustness}

\begin{figure}
	\centering
	\includegraphics[width=\linewidth]{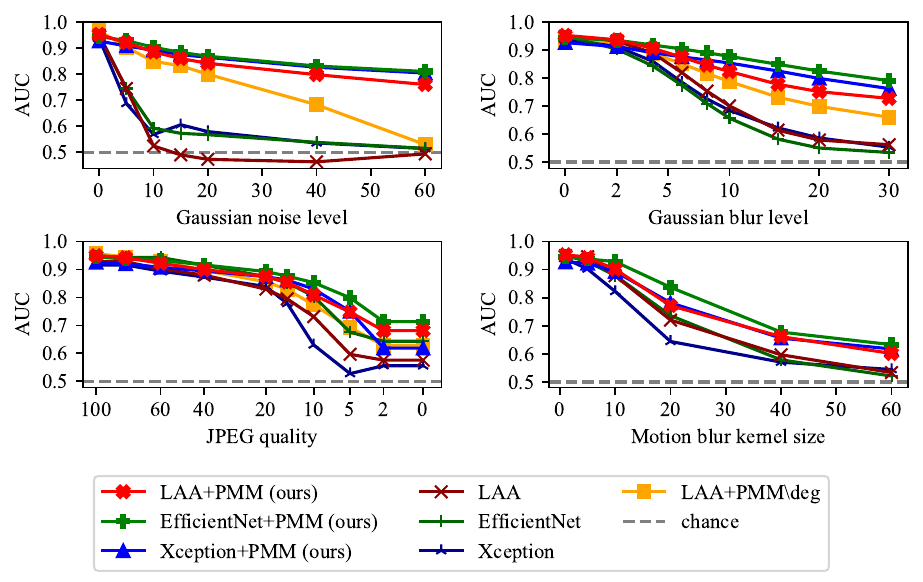}
	\caption{\textbf{Robustness to image degradations} of s-o-t-a deepfake detectors LAA~\protect\cite{nguyen2024laa} and SBI (EfficientNet and Xception backbones)~\protect\cite{Shiohara2022SBI} vs. ours, tested on the test-split of FF++~\protect\cite{roessler2019faceforensicspp}. Note that for the low-quality settings, our methods consistently outperform the s-o-t-a. In particular, for Gaussian noise and blur, we outperform the baselines by $25\%$ and $17\%$ AUC, respectively. PMM$\setminus$deg refers to a variant of PMM, trained without the specific test degradation. Motion blur is generally never in the training.}
	\label{fig:robustness}
\end{figure}
\begin{figure*}
	\centering
	\begin{subfigure}[t]{0.24\linewidth}
		\centering
		\includegraphics[width=1\linewidth]{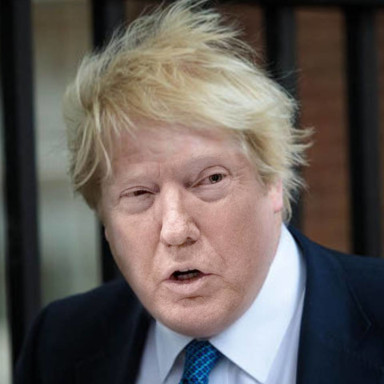}
		\caption{\textcolor{red}{Fake} image}
		\label{fig:johnson_normal}
	\end{subfigure}
	\begin{subfigure}[t]{0.24\linewidth}
		\centering
		\includegraphics[width=1\linewidth]{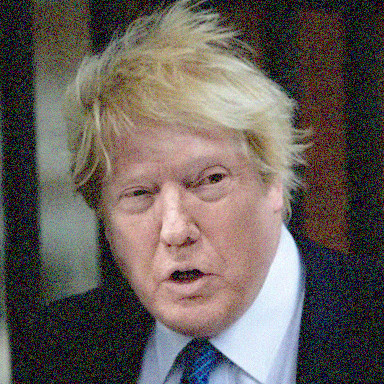}
		\caption{\textcolor{red}{Fake} image + Gaussian noise}
		\label{fig:johnson_noise}
	\end{subfigure}
	\begin{subfigure}[t]{0.24\linewidth}
		\centering
		\includegraphics[width=1\linewidth]{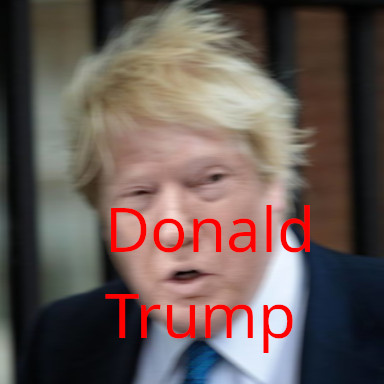}
		\caption{\textcolor{red}{Fake} image + motion blur + text}
		\label{fig:johnson_blur}
	\end{subfigure}
	\begin{subfigure}[t]{0.24\linewidth}
		\centering
		\includegraphics[width=1\linewidth]{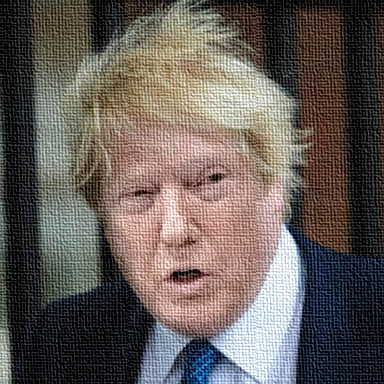}
		\caption{\textcolor{red}{Fake} image + canvas effect}
		\label{fig:johnson_canvas}
	\end{subfigure}
	\caption{Donald Trump's face swapped onto Boris Johnson~\protect\cite{faceswaponline2020trumpjohnson}. 
		(a) Both LAA and ours easily recognize it as \textcolor{red}{fake}. However, several image degradations (b)-(d) lead LAA to predict \textcolor{green}{real}, whereas ours continues to recognize the image as \textcolor{red}{fake}. Note that neither motion blur nor a canvas effect was part of the training degradations, and still, our model is robust to these changes.}
	\label{fig:johnson}
\end{figure*}

To assess the robustness to image degradations of the s-o-t-a SBI~\cite{Shiohara2022SBI} (EfficientNet~\cite{Tan2019EfficientNetRM} and Xception~\cite{Chollet2016XceptionDL} backbones) and LAA~\cite{nguyen2024laa} deepfake detectors and ours (+PMM), we apply different degradations to the test split of FaceForensics++~\cite{roessler2019faceforensicspp} and report AUC.
We chose the following degradations: Gaussian noise with $\sigma\in[0..60]$, Gaussian blur with $\sigma\in[0..30]$, JPEG compression with Quality Factors $\in[0..100]$ and motion blur with kernel size $\in[0..60]$.
 This is an in-dataset evaluation to provide the models with the best-case scenario and only measure the effects of image degradations. For every image, we apply the described degradation before passing it to the models. \Cref{fig:robustness} shows the robustness results. Our method consistently outperforms the baselines at the strong degradation settings. Additional degradations are shown in the supplementary.

\vspace{2mm}\noindent\textbf{Gaussian Noise.}
Gaussian noise negatively affects all baselines, even for low amounts of noise ($\sigma < 10$). At $\sigma = 15$, LAA performs worse than chance ($0.49$ AUC). Both versions of SBI are slightly more robust at about $0.6$ AUC, whereas the PMM versions perform significantly better at $\geq 0.86$ AUC. Even for the strongest tested setting of $\sigma = 60$, our models still perform at $\geq 0.76$ AUC, whereas the baselines perform at chance level.

\vspace{2mm}\noindent\textbf{Gaussian Blur.}
At $\sigma = 6$, LAA, the best of the baselines, drops by $13\%$, whereas the worst of our methods (LAA+PMM) only drops by $7.8\%$, showing the baselines' lack of robustness, even for small amounts of blur. This difference increases for larger values of $\sigma$. At $\sigma = 30$, the difference between the best baseline and the worst of ours is $16.5\%$, with the baselines performing close to chance level.

\vspace{2mm}\noindent\textbf{JPEG Compression.}
For JPEG compression, we see a clear difference between the backbones: EfficientNet is best, followed by LAA and Xception. At low quality levels ($<40$), PMM can improve all backbones significantly. LAA+PMM, for example, outperforms LAA by $10.6\%$ at the lowest quality setting.

\vspace{2mm}\noindent\textbf{Motion blur.}
We implement motion blur using an average blur kernel $k_{ij} = \frac{1}{w}$ of size $1 \times w$. This case was not part of the training data. Still, PMM consistently outperforms the baselines. At a kernel size of $w = 60$, LAA+PMM outperforms LAA by $6.8\%$ AUC. This shows the generalization ability of our model to unseen degradation types.

\vspace{2mm}\noindent\textbf{Leave-one-out training.}
We also test models that used the entire PMM model, except for the specific test degradation, during training. As expected, they perform worse than the full PMM model, but better than LAA, indicating generalization to unseen degradations (see supplementary).

\vspace{2mm}\noindent\textbf{New dataset.}
To test robustness to newer methods, we compare LAA vs. LAA+PMM on the FS and FR parts of DF40~\cite{yan2024df40} based on CDF data. LAA+PMM outperforms LAA with 0.644 vs. 0.636 AUC, indicating that PMM can also improve generalization to newer methods. More details are in the supplementary.

\subsection{Qualitative demonstration}
\label{sec:qualitative_demonstration}

\Cref{fig:johnson_normal} shows a deepfake of Donald Trump's face placed onto the head of Boris Johnson. Both LAA and our method can easily detect the image as fake with high certainty. The decision is likely based on the difference in color between the face and the neck. However, once we add a little bit of Gaussian noise (\cref{fig:johnson_noise}), the prediction of LAA flips completely. While we only lose some certainty in the image being fake, LAA is now very certain that the image is real.

\Cref{fig:johnson_blur} shows the same fake image with applied motion blur and overlayed text. Again, the prediction for LAA flips, whereas ours remains "fake". \Cref{fig:johnson_canvas} has a canvas effect applied to the image, which also causes the LAA prediction to change to "real". Note that both motion blur and the canvas effect were not part of the training data.

This is a serious problem, as the noisy image is still easily recognizable by a human. So, together with some made-up explanation of why the image is noisy, a faker would now have a completely fake image, which is clearly and with a high degree of certainty labeled as real by a state-of-the-art detector. The same can be said about the other two images. Motion blur is easy to explain, and text on an image is common. 
While the canvas effect is certainly strange, it highlights how vulnerable the baseline is to degradations. An attacker has a lot of freedom to choose some degradation that has the desired effect.
\textit{For this reason, we, again, want to stress the importance of robustness.}

	\section{Limitations}
\label{sec:limitations}

Although our performance compared to the state-of-the-art is very strong, we want to acknowledge some limitations of our model. While we successfully address image processing degradations, we do not cover robustness towards adversarial attacks \cite{Szegedy2013AdversarialAttacks}. Therefore, adversarially constructed disturbances may still fool our detector. Moreover, we lose some performance on the high-quality dataset CDFv2, where the additional robustness is not beneficial, and our more general detector performs slightly worse than the specialized high-quality baselines. Possible solutions could be training a larger backbone or detecting high-quality images and using one of the original models in that case.
	\section{Conclusion}
\label{sec:conclusion}

In this paper, we proposed a more general approach for generating pseudo-fakes. We utilize more diverse masks, blending techniques, distractors, generator artifacts, and image degradations to avoid a situation where a malicious deepfaker can easily avoid detection by only a partial face swap or some kind of image degradation. Extensive experiments show the validity of our approach as well as strong improvements on popular benchmark datasets, even without adding degradations during test time.
Given that our Practical Manipulation Model is not bound to any specific architecture, we propose to consider adding our method to future works in order to increase their performance and robustness, especially towards maliciously hidden deepfakes.
	
	\clearpage
	{
		\small
		\bibliographystyle{ieeenat_fullname}
		\bibliography{main}
	}
	\clearpage
\maketitlesupplementary
\setcounter{section}{0}
\setcounter{figure}{0}
\setcounter{table}{0}
\appendix

\renewcommand{\thetable}{\Alph{table}}
\renewcommand{\thefigure}{\Alph{figure}}
\renewcommand{\thesection}{\Alph{section}}

\section*{Overview}

This document provides additional information omitted from the main paper for brevity. \Cref{sec:assassination_image} explains the inference process on the images in \cref{fig:teaser}. \Cref{sec:more_example_images} provides additional example images of our Practical Manipulation Model, in the same way as in \cref{tab:pmm_examples}. Additional details for our method are described in \cref{sec:method_details}, whereas \cref{sec:full_ablation} provides more details on the ablation study. \Cref{sec:additional_robustness} provides additional robustness examples, similar to \cref{sec:robustness} in the main paper and \cref{sec:gradcam} qualitatively evaluates robustness using GradCAM \cite{Selvaraju2016GradCAMVE}. Finally, \cref{sec:qualitative} provides more qualitative examples of our model and LAA \cite{nguyen2024laa}, while \cref{sec:recognizability} presents a metric for the hardness of our degradations and \cref{sec:implementation} describes some details of our implementation.

\section{Inference on the assassination attempt image}
\label{sec:assassination_image}

\Cref{fig:teaser} shows a real image \cite{moreechampion2024trump_asassination} and its corresponding real-world deepfake \cite{dfrlab2024trump_asassination}, which can avoid detection due to being slightly blurry. In the image, the faces of the two Secret Service agents have been modified to make them smile. 

We tested the detection on the agent on the right because he is more obviously visible, properly exposed, and looking at the camera. Since both detectors are meant to be applied to crops of faces, we manually created a crop around the agent's head and performed detection on that.

\section{More example images}
\label{sec:more_example_images}

\begin{figure}
    \centering
    \newcommand{\pmmExampleScale}{0.12}
    \begin{tabular}{|cc|cc|}
    \hline
        \multicolumn{2}{|c|}{LAA~\cite{nguyen2024laa}} & \multicolumn{2}{c|}{ours} \\ \hline
        real & fake & real & fake \\
         \includegraphics[scale=\pmmExampleScale]{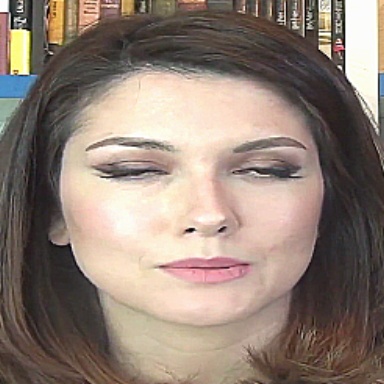} &
         \includegraphics[scale=\pmmExampleScale]{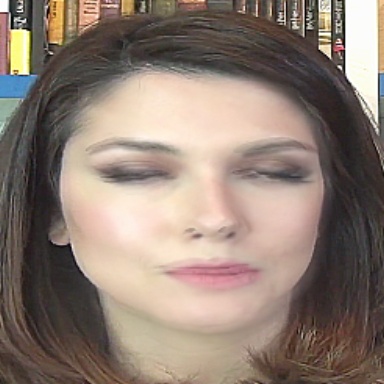} &
         \includegraphics[scale=\pmmExampleScale]{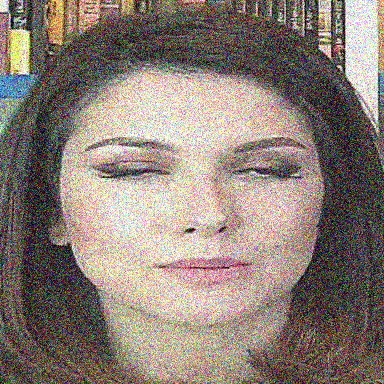} & 
         \includegraphics[scale=\pmmExampleScale]{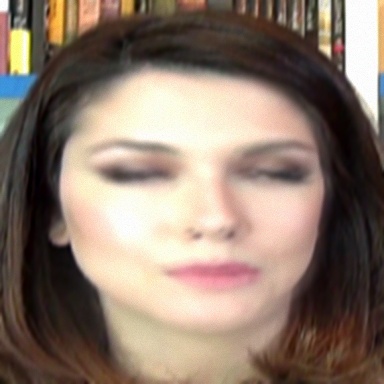} \\

         \includegraphics[scale=\pmmExampleScale]{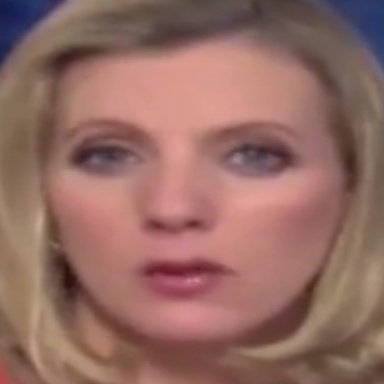} &
         \includegraphics[scale=\pmmExampleScale]{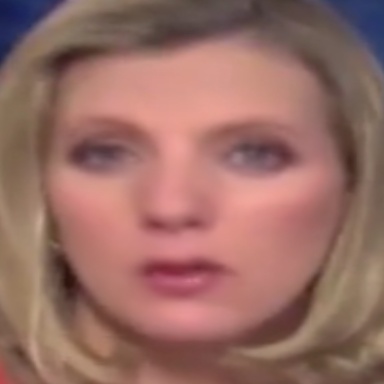} &
         \includegraphics[scale=\pmmExampleScale]{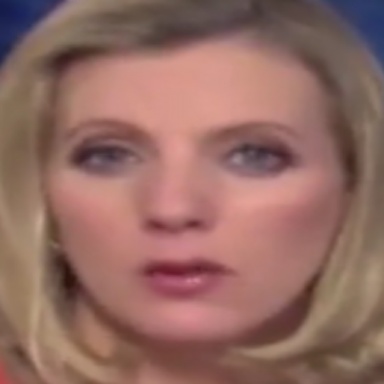} & 
         \includegraphics[scale=\pmmExampleScale]{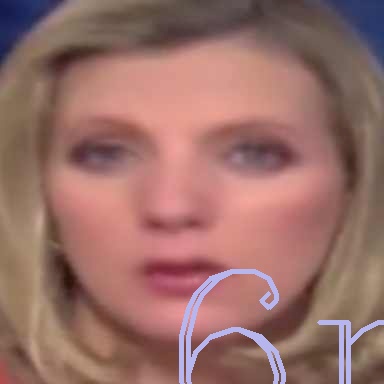} \\

         \includegraphics[scale=\pmmExampleScale]{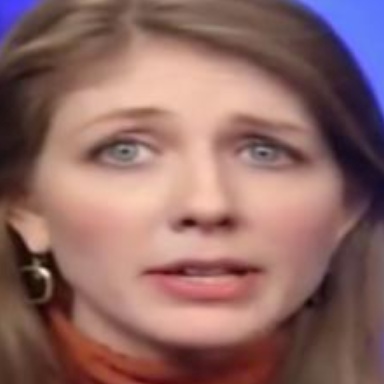} &
         \includegraphics[scale=\pmmExampleScale]{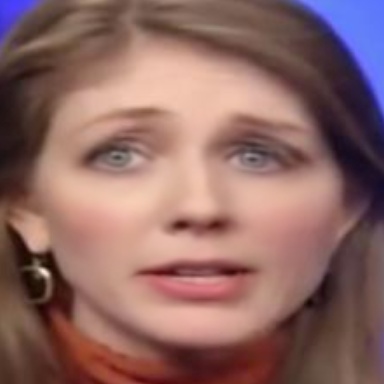} &
         \includegraphics[scale=\pmmExampleScale]{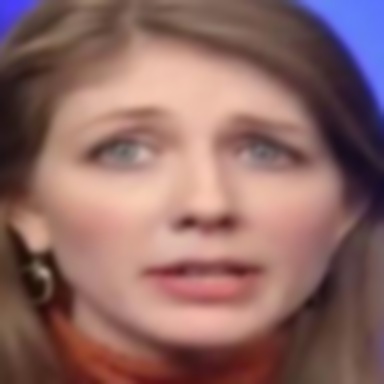} & 
         \includegraphics[scale=\pmmExampleScale]{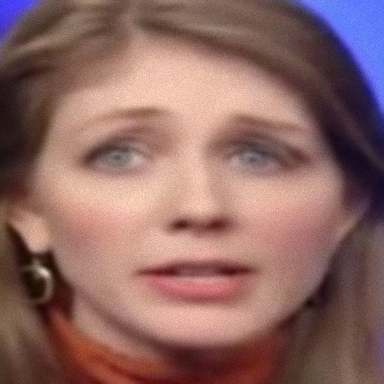} \\

         \includegraphics[scale=\pmmExampleScale]{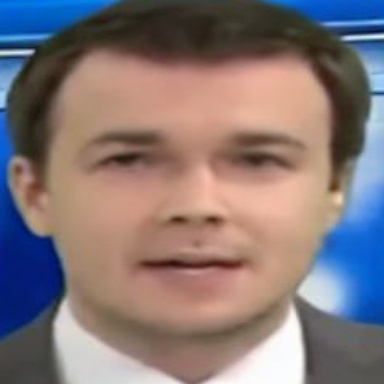} &
         \includegraphics[scale=\pmmExampleScale]{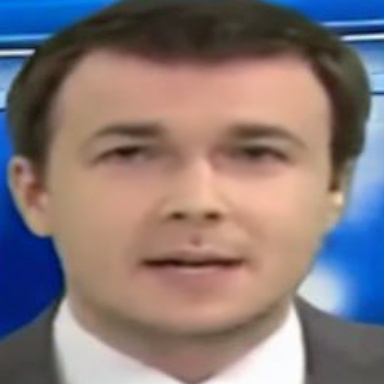} &
         \includegraphics[scale=\pmmExampleScale]{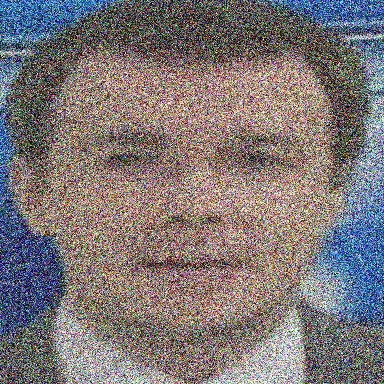} & 
         \includegraphics[scale=\pmmExampleScale]{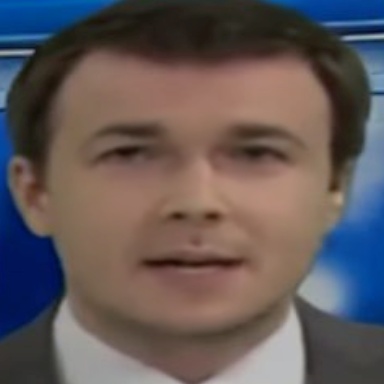} \\

         \includegraphics[scale=\pmmExampleScale]{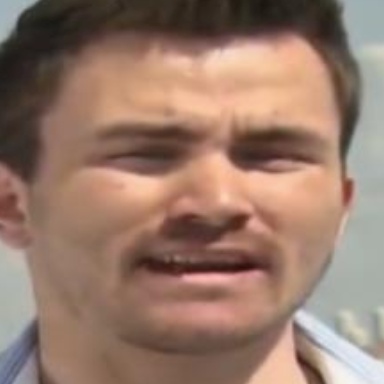} &
         \includegraphics[scale=\pmmExampleScale]{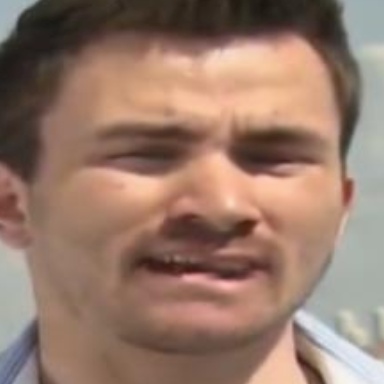} &
         \includegraphics[scale=\pmmExampleScale]{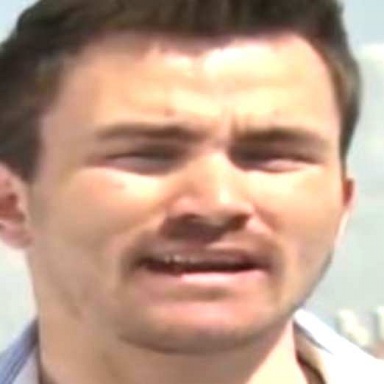} & 
         \includegraphics[scale=\pmmExampleScale]{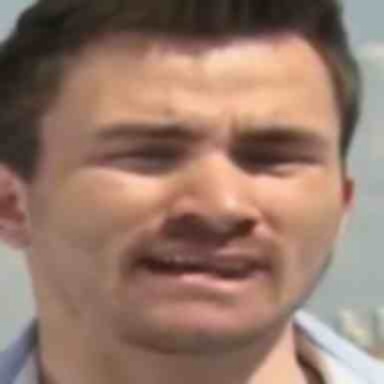} \\

         \includegraphics[scale=\pmmExampleScale]{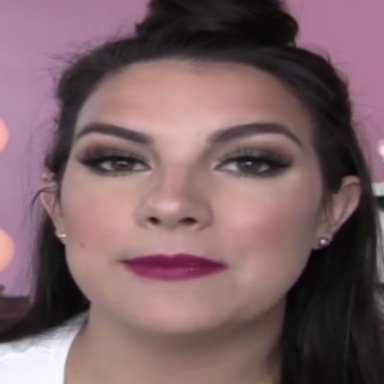} &
         \includegraphics[scale=\pmmExampleScale]{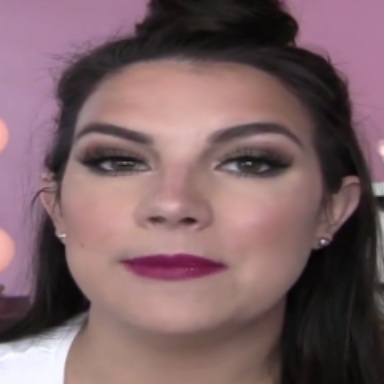} &
         \includegraphics[scale=\pmmExampleScale]{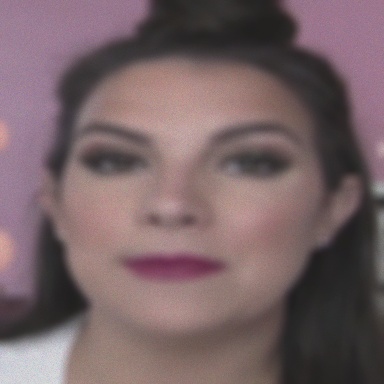} & 
         \includegraphics[scale=\pmmExampleScale]{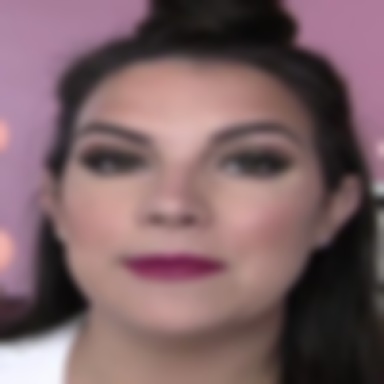} \\

         \includegraphics[scale=\pmmExampleScale]{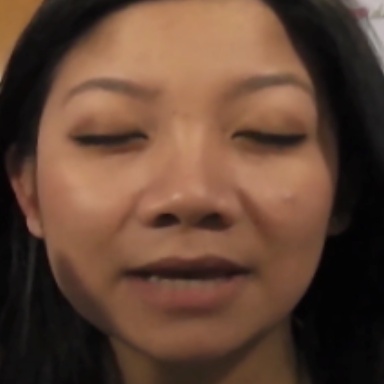} &
         \includegraphics[scale=\pmmExampleScale]{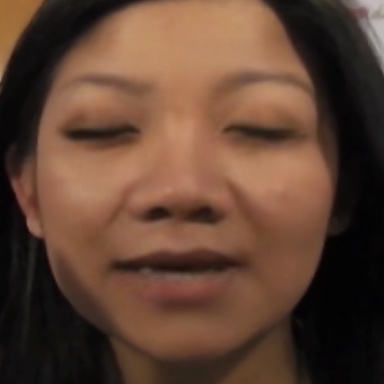} &
         \includegraphics[scale=\pmmExampleScale]{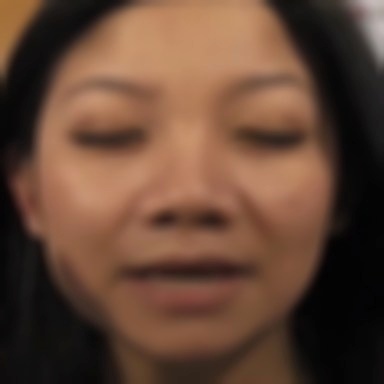} & 
         \includegraphics[scale=\pmmExampleScale]{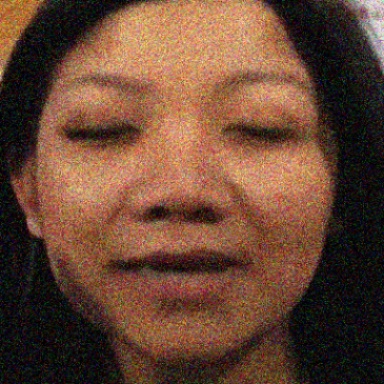} \\

         \includegraphics[scale=\pmmExampleScale]{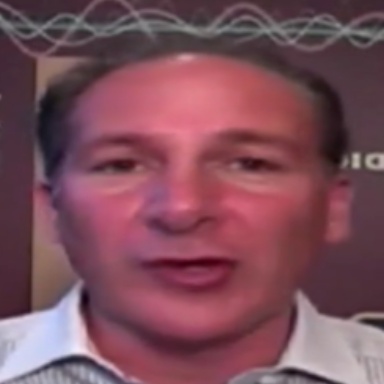} &
         \includegraphics[scale=\pmmExampleScale]{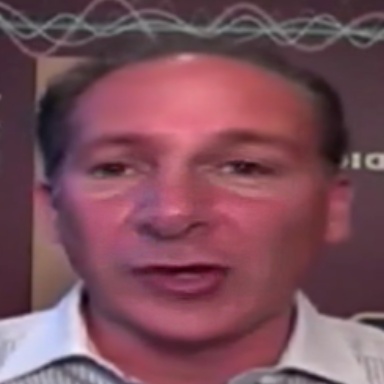} &
         \includegraphics[scale=\pmmExampleScale]{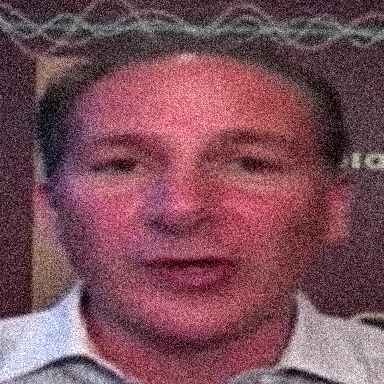} & 
         \includegraphics[scale=\pmmExampleScale]{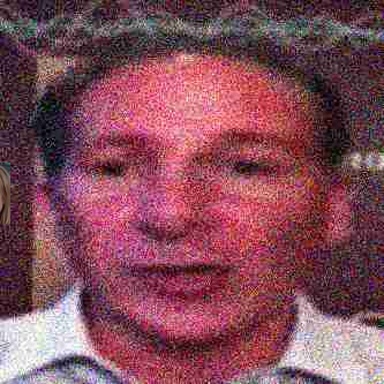} \\
         \hline
    \end{tabular}
    \caption{Additional example images from our method and LAA~\protect\cite{nguyen2024laa}. This is an extension of \cref{tab:pmm_examples}. The original images are taken from the FaceForensics++ dataset~\protect\cite{roessler2019faceforensicspp}.}
    \label{tab:pmm_examples_supp}
\end{figure}

\Cref{tab:pmm_examples_supp} shows some additional training images. The table follows the same concept as \cref{tab:pmm_examples} in the main paper, providing more examples that could not be included for space reasons.

\begin{table*}
    \centering
    \newcommand{\pmmExampleScale}{0.11}
    \begin{tblr}{
    colspec={cX[16mm,h]X[16mm,h]X[16mm,h]X[16mm,h]X[16mm,h]X[16mm,h]X[16mm,h]},
    stretch = 0,
    rowsep = 2pt,
    colsep=1pt
    }
        \hline
        Variant & \SetCell[c=6]{c} example training images \\ \hline
        Original (LAA) & 
        \includegraphics[scale=\pmmExampleScale]{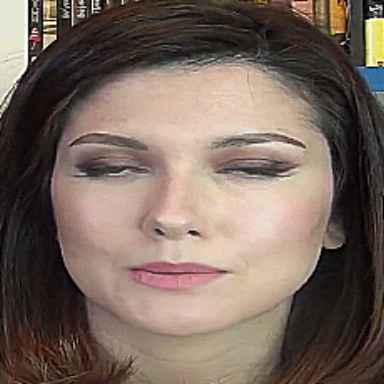} & 
        \includegraphics[scale=\pmmExampleScale]{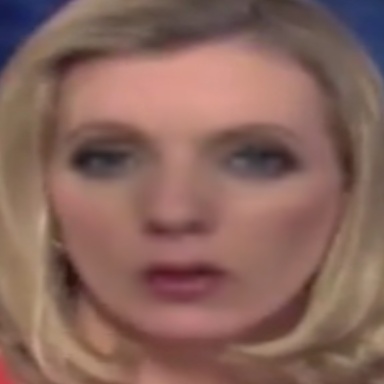} & 
        \includegraphics[scale=\pmmExampleScale]{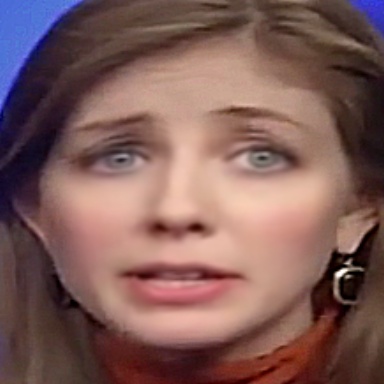} & 
        \includegraphics[scale=\pmmExampleScale]{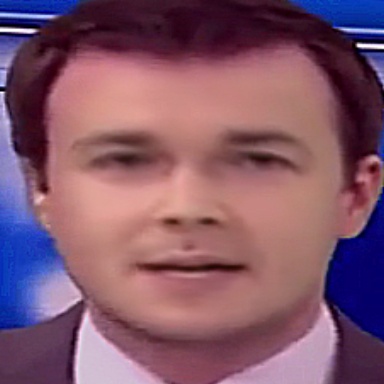}  & 
        \includegraphics[scale=\pmmExampleScale]{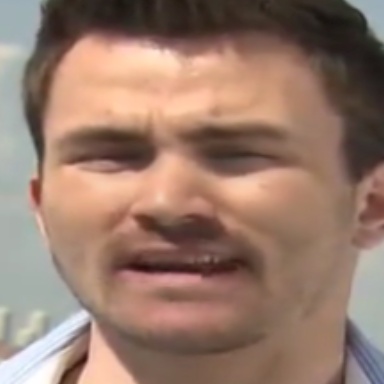}  & 
        \includegraphics[scale=\pmmExampleScale]{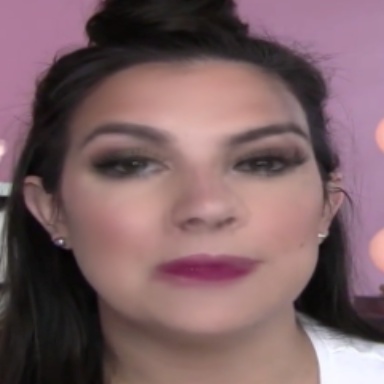}  & 
        \includegraphics[scale=\pmmExampleScale]{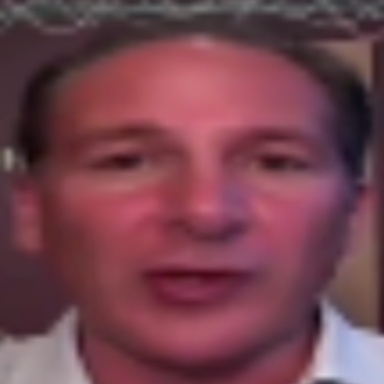} 
        \\
        + Degradations & 
        \includegraphics[scale=\pmmExampleScale]{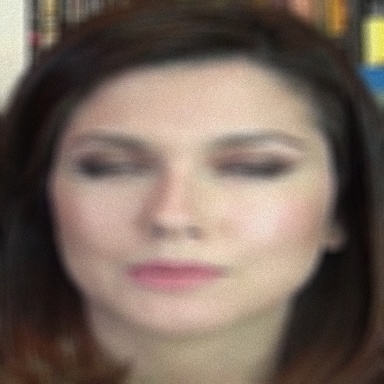} & 
        \includegraphics[scale=\pmmExampleScale]{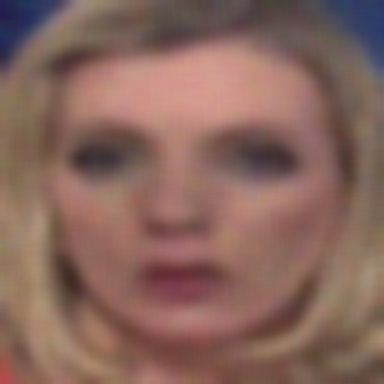} & 
        \includegraphics[scale=\pmmExampleScale]{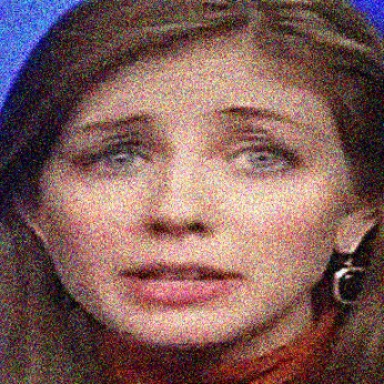} & 
        \includegraphics[scale=\pmmExampleScale]{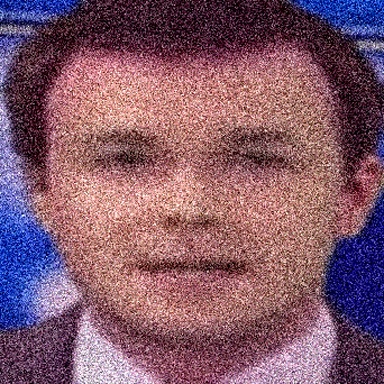} & 
        \includegraphics[scale=\pmmExampleScale]{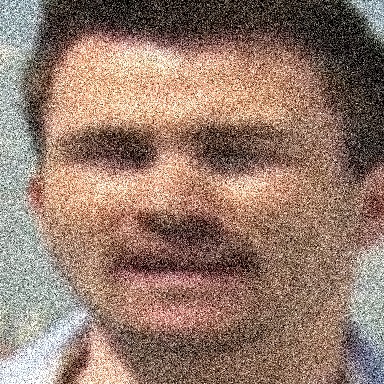} & 
        \includegraphics[scale=\pmmExampleScale]{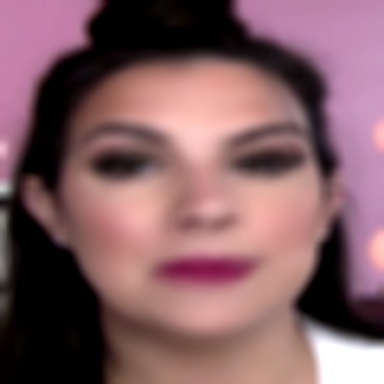} & 
        \includegraphics[scale=\pmmExampleScale]{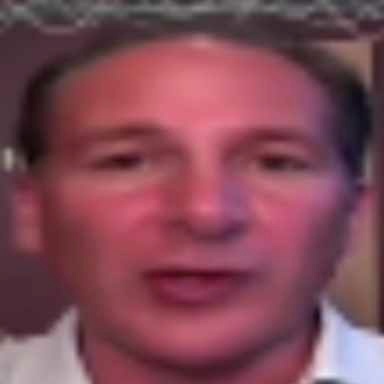} \\
        + Additional Masks & 
        \includegraphics[scale=\pmmExampleScale]{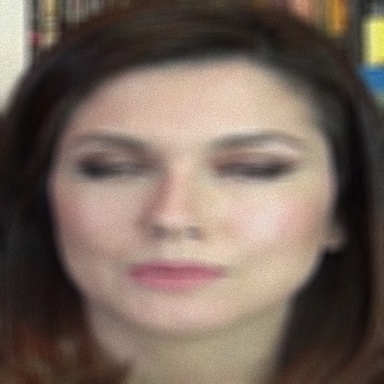} &
        \includegraphics[scale=\pmmExampleScale]{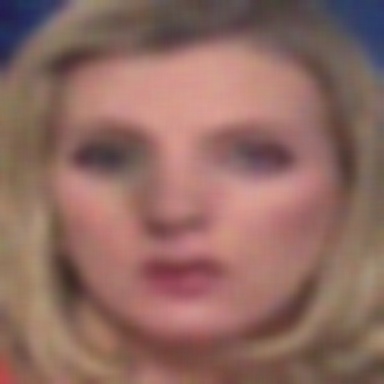} &
        \includegraphics[scale=\pmmExampleScale]{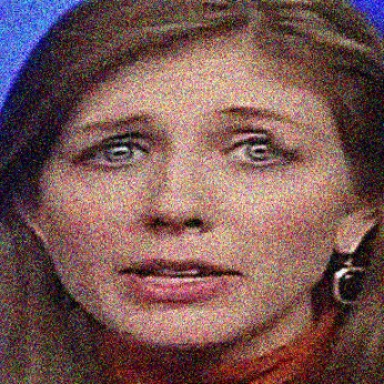} &
        \includegraphics[scale=\pmmExampleScale]{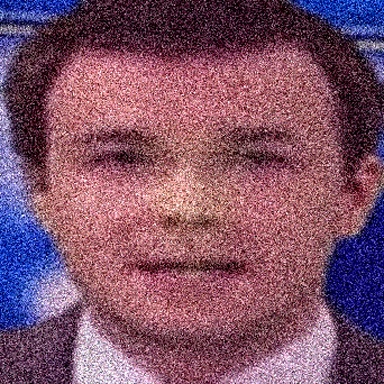} &
        \includegraphics[scale=\pmmExampleScale]{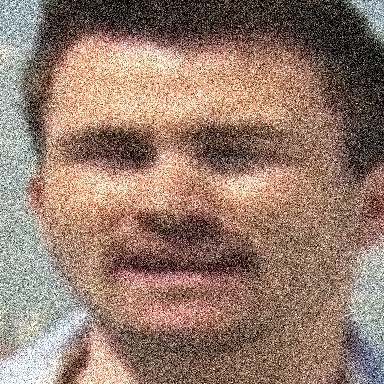} &
        \includegraphics[scale=\pmmExampleScale]{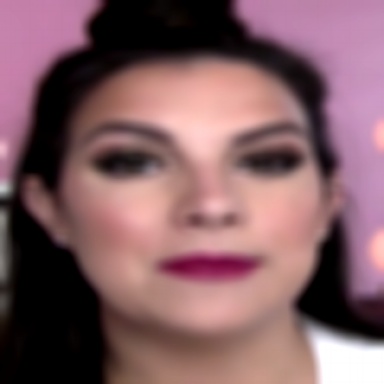} &
        \includegraphics[scale=\pmmExampleScale]{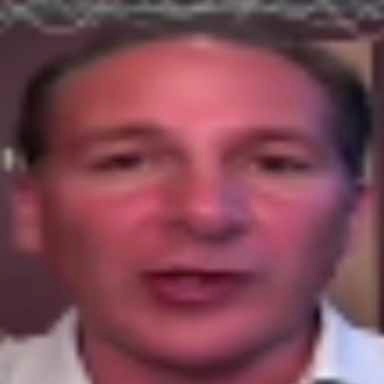} \\
        + Poisson blending & 
        \includegraphics[scale=\pmmExampleScale]{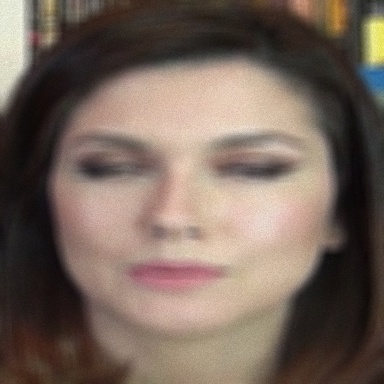} & 
        \includegraphics[scale=\pmmExampleScale]{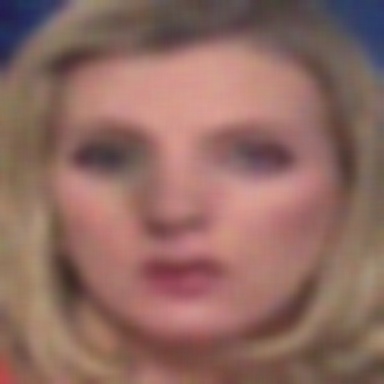} & 
        \includegraphics[scale=\pmmExampleScale]{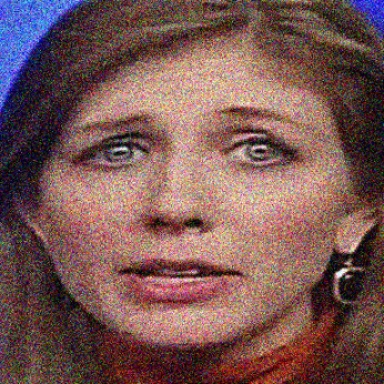} & 
        \includegraphics[scale=\pmmExampleScale]{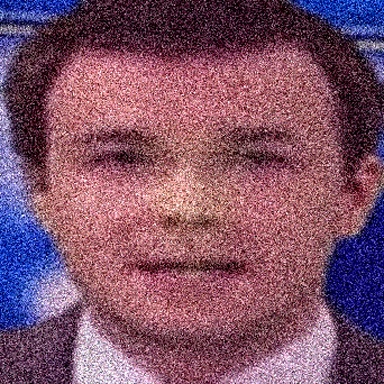} & 
        \includegraphics[scale=\pmmExampleScale]{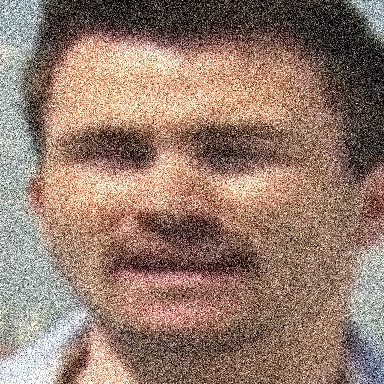} & 
        \includegraphics[scale=\pmmExampleScale]{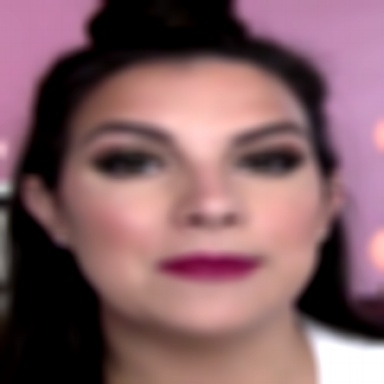}& 
        \includegraphics[scale=\pmmExampleScale]{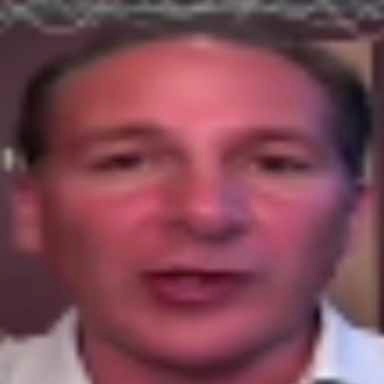} \\
        + Distractors & 
        \includegraphics[scale=\pmmExampleScale]{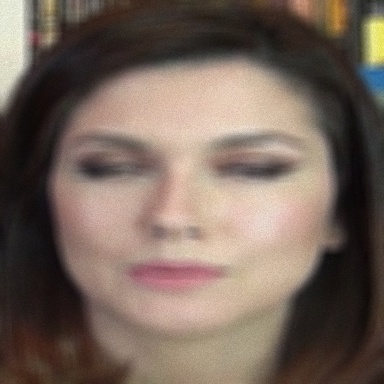} & 
        \includegraphics[scale=\pmmExampleScale]{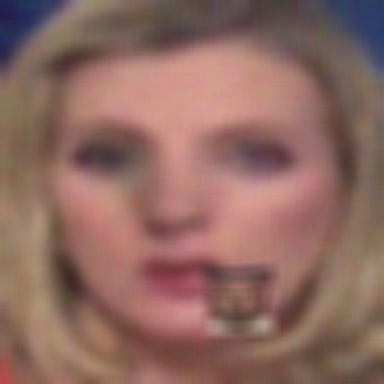} & 
        \includegraphics[scale=\pmmExampleScale]{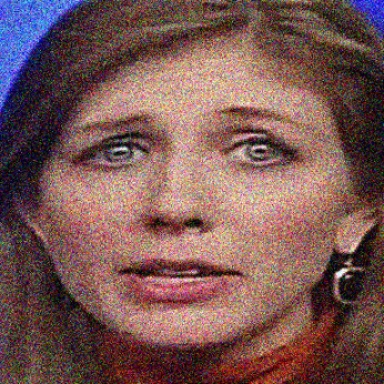} & 
        \includegraphics[scale=\pmmExampleScale]{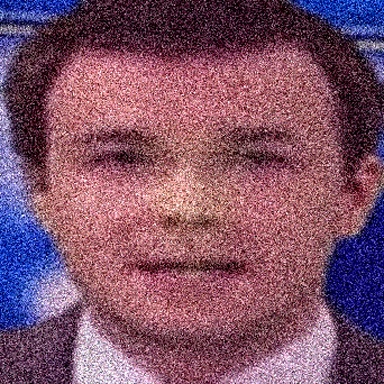} & 
        \includegraphics[scale=\pmmExampleScale]{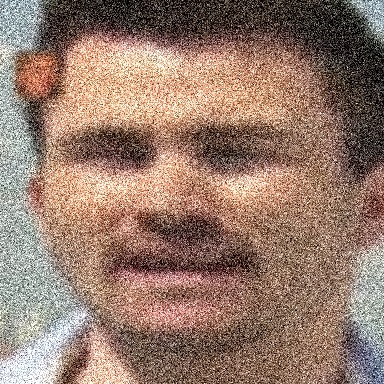} & 
        \includegraphics[scale=\pmmExampleScale]{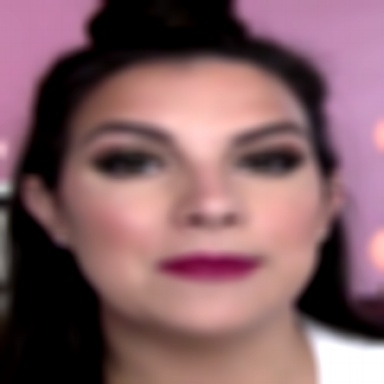} & 
        \includegraphics[scale=\pmmExampleScale]{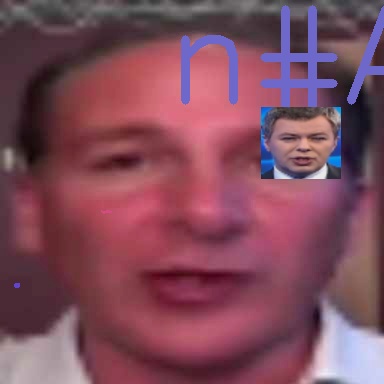} \\
        + Generator Artifacts (PMM) & 
        \includegraphics[scale=\pmmExampleScale]{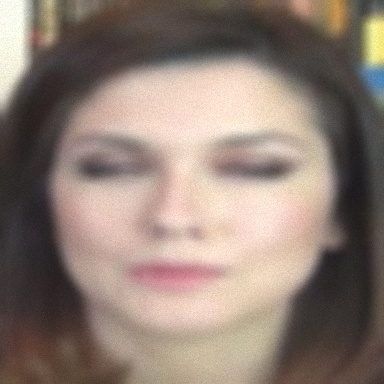} & 
        \includegraphics[scale=\pmmExampleScale]{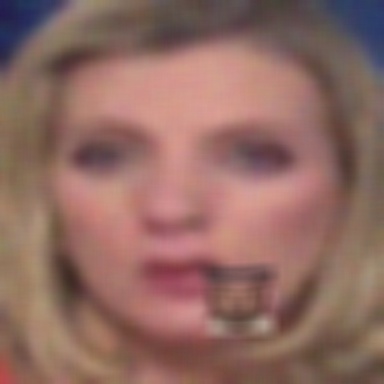} & 
        \includegraphics[scale=\pmmExampleScale]{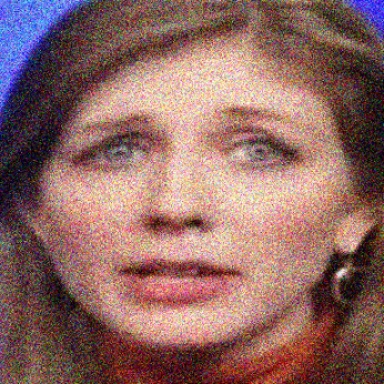} & 
        \includegraphics[scale=\pmmExampleScale]{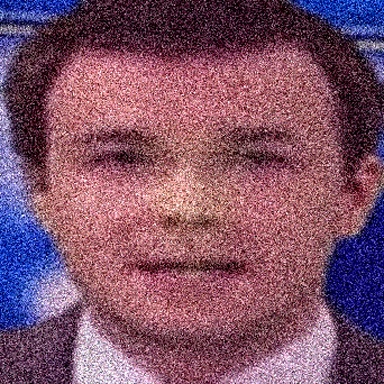} & 
        \includegraphics[scale=\pmmExampleScale]{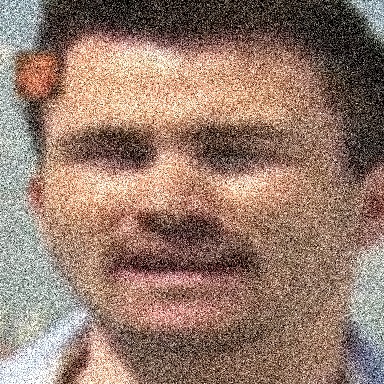} & 
        \includegraphics[scale=\pmmExampleScale]{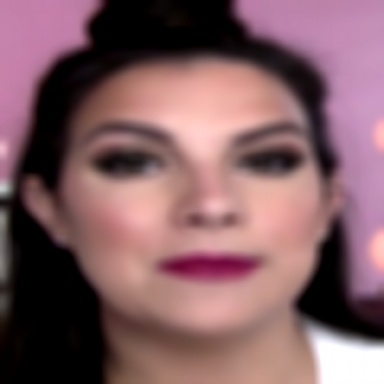} & 
        \includegraphics[scale=\pmmExampleScale]{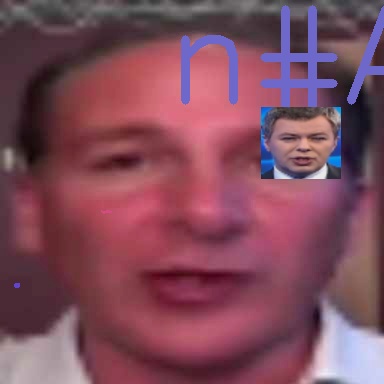} \\
    \end{tblr}
    \caption{\textbf{Sample images during our ablation study.} The images are examples of the kind of fake data the model sees during the ablation. Note that not all changes are visible to the human eye. Images are selected to emphasize the effect of the visible steps, avoiding cases where a change is randomly not used. Original images are taken from FaceForensics++~\protect\cite{roessler2019faceforensicspp}.}
    \label{tab:ablation_examples}
\end{table*}

\section{Method details}
\label{sec:method_details}

This section provides explanations of two details of our method, which were omitted from the paper due to space constraints.
\subsection{Noise}
 Channel-wise correlated Gaussian noise is given by \begin{equation}
\Sigma' = \left| \left(l_2' \cdot s\right)^2 \cdot U^TDU \right|
\end{equation} for some orthogonal matrix $U$ and diagonal matrix $D$. The diagonal elements of $D$ are randomly sampled from $\mathit{Uniform}(0, 1)$. $U$ is entirely randomly sampled from $\mathit{Uniform}(0, 1)$ and then orthogonalized. $l_2' = \frac{l_2}{255}$ is the noise level in range $[0, 1]$. When calculating in range $\{0, 1, ..., 255\}$, the result has to be scaled back up: $\Sigma = 255 \cdot \Sigma'$

\subsection{Distractors}
\label{sec:distractor_details}
\paragraph*{Multiple distractors per image.}
As described in the paper, multiple distractors may be added to a single image. The number of distractors is geometrically distributed. This means they are added until the random decision with probability $p_d$ fails. Alternatively, we limit the number of distractors to 10 to avoid cluttering the image too much for large values of $p_d$. However, since we use $p_d = 0.2$, this case is only expected to happen with a probability of $0.2^{11} \approx 2 \cdot 10^{-8}$ and should, therefore, not have much effect in our experiments.
\paragraph*{Text settings.}
The following settings are chosen for the overlaid text: \begin{enumerate}
    \item The text itself is a concatenation of $n \sim \mathit{Uniform}(0, 10)$ randomly selected characters, chosen from the list of all printable characters.
    \item The text is placed at position (lower left corner) $x \sim \mathit{Uniform}(-100, \mathit{WIDTH})$ and $y \sim \mathit{Uniform}(0, HEIGHT+100)$. Positions outside of the image region result in the text being partially visible.
    \item Font face is uniformly chosen from $\{0, ..., 7\}$.
    \item Font scale is chosen from $\mathit{Uniform}(0, 8)$.
    \item Color is uniformly chosen from $\{0, ..., 255\}^3$.
    \item Line thickness is uniformly chosen from $\{1, 2, ..., 8\}$.
    \item Line type is uniformly chosen from $\{0, 1, 2\}$.
\end{enumerate}

\section{Details on the ablation study}
\label{sec:full_ablation}

\Cref{tab:ablation_examples} shows examples of the types of images our model sees while training each of the ablation study variants. The images are selected to show the change of each variant (where possible), so the distribution is skewed towards our changes and away from the baseline. However, all images shown can appear during training.

\vspace{2mm}\noindent\textbf{Parameters.} During training, we use degradations with $p=50\%$ and strength $s=0.5$. Poisson blending is used $p_p=50\%$ of the time, and distractors are placed in $p_d = 20\%$ of the images, as described in \cref{sec:distractor_details}. Finally, we use both types of Generator Artifacts in $p_g = 25\%$ of the images, each. 

From our testing, the model is relatively robust to hyperparameter choice, except for $p \approx 1$, where clean images appear too rarely. Especially small values of $p$ and $s$ have never been observed to hurt performance (\eg $+3.42\%$ AUC on DFDCP for $p = s = 0.3$).

\begin{table*}
    \centering
    \newcommand{\pmmExampleScale}{0.12}
    \begin{tblr}{
    colspec={ccX[16.5mm,h]X[16.5mm,h]X[16.5mm,h]X[16.5mm,h]X[16.5mm,h]},
    stretch = 0,
    rowsep = 2pt,
    colsep=2pt
    }
        \hline
        Deepfake- & Gaussian Noise & \SetCell{c} real & \SetCell[c=4]{c} fake \\
        detector & $\sigma = 20$ & \SetCell{c} Original & \SetCell{c} DF \cite{deepfakes} & \SetCell{c} FS \cite{faceswap} & \SetCell{c} F2F \cite{Thies2016Face2Face} & \SetCell{c} NT \cite{Thies2019NeuralTextures} \\ \hline
        SBI \cite{Shiohara2022SBI} & & 
        \includegraphics[scale=\pmmExampleScale]{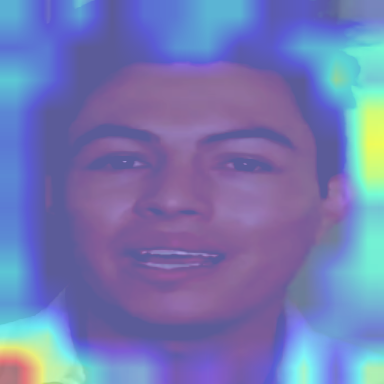} & 
        \includegraphics[scale=\pmmExampleScale]{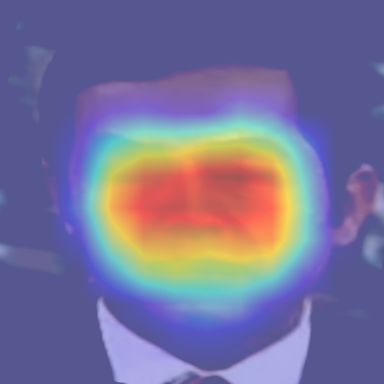} & 
        \includegraphics[scale=\pmmExampleScale]{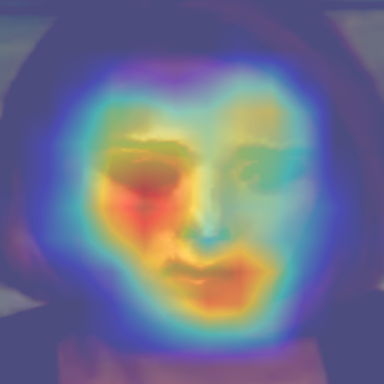} & 
        \includegraphics[scale=\pmmExampleScale]{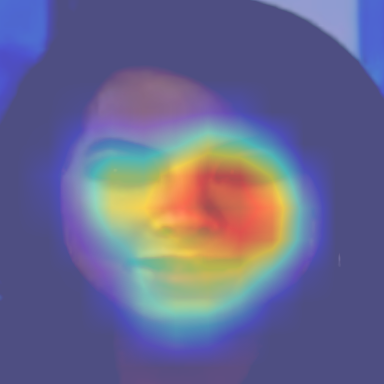} & 
        \includegraphics[scale=\pmmExampleScale]{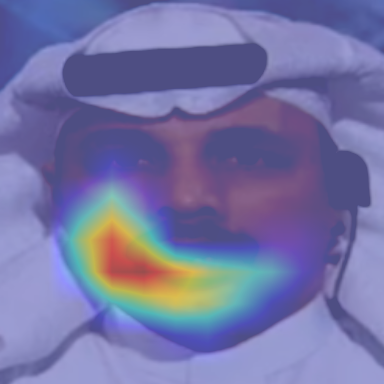} \\
        SBI \cite{Shiohara2022SBI} & \checkmark & 
        \includegraphics[scale=\pmmExampleScale]{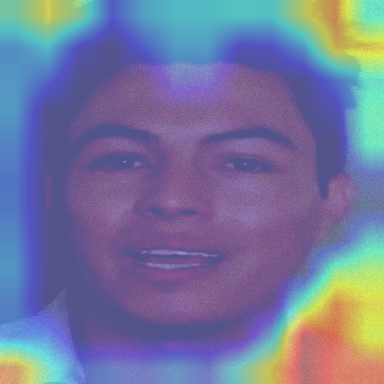} & 
        \includegraphics[scale=\pmmExampleScale]{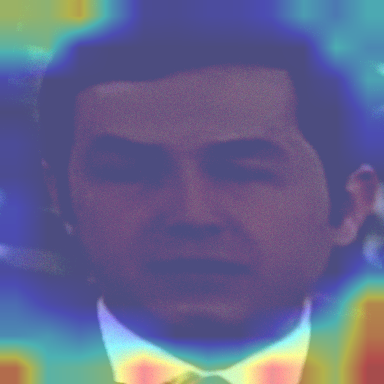} & 
        \includegraphics[scale=\pmmExampleScale]{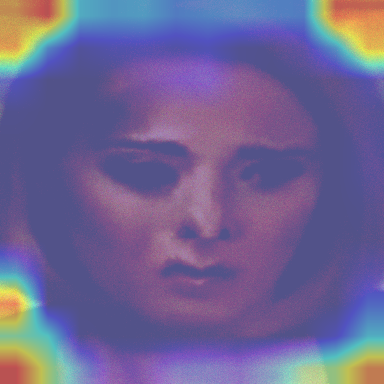} & 
        \includegraphics[scale=\pmmExampleScale]{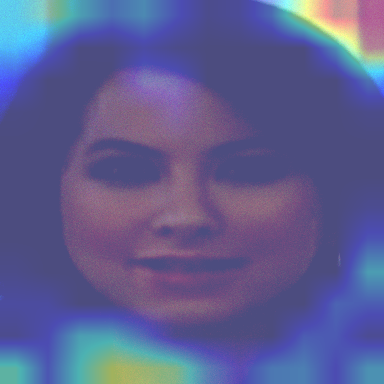} & 
        \includegraphics[scale=\pmmExampleScale]{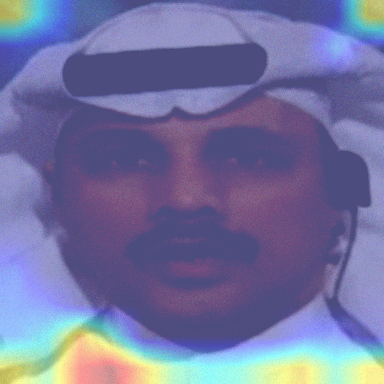} \\ \hline
        LAA \cite{nguyen2024laa} & & 
        \includegraphics[scale=\pmmExampleScale]{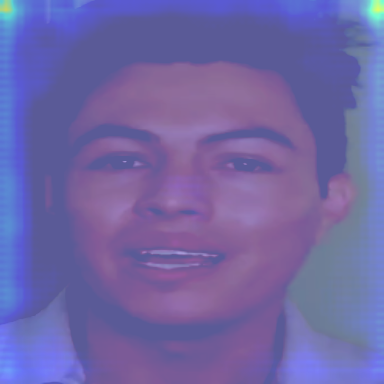} & 
        \includegraphics[scale=\pmmExampleScale]{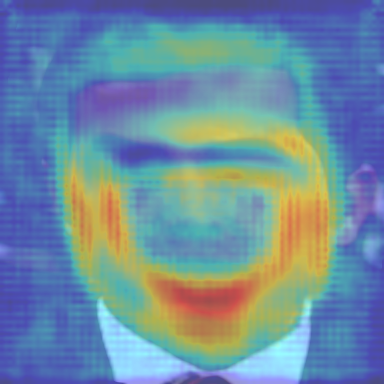} & 
        \includegraphics[scale=\pmmExampleScale]{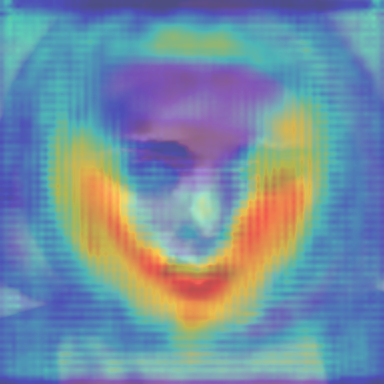} & 
        \includegraphics[scale=\pmmExampleScale]{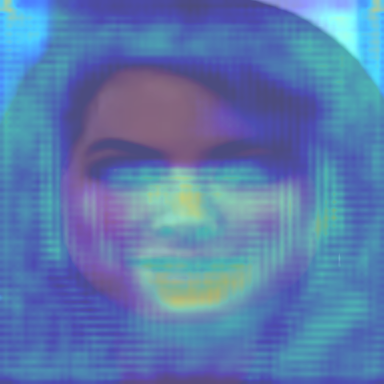} & 
        \includegraphics[scale=\pmmExampleScale]{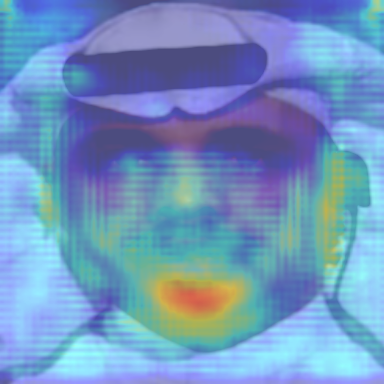} \\
        LAA \cite{nguyen2024laa} & \checkmark & 
        \includegraphics[scale=\pmmExampleScale]{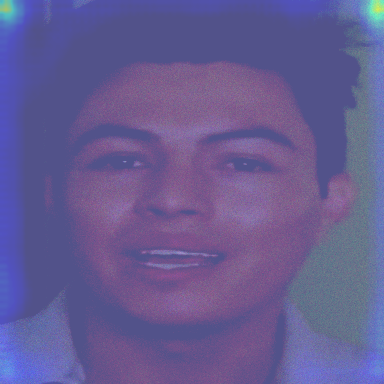} & 
        \includegraphics[scale=\pmmExampleScale]{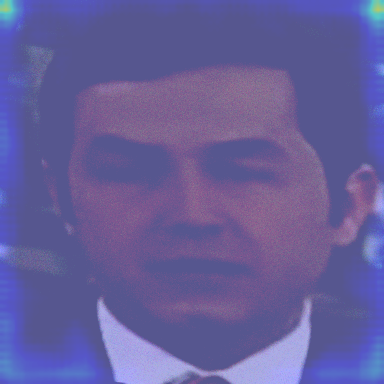} & 
        \includegraphics[scale=\pmmExampleScale]{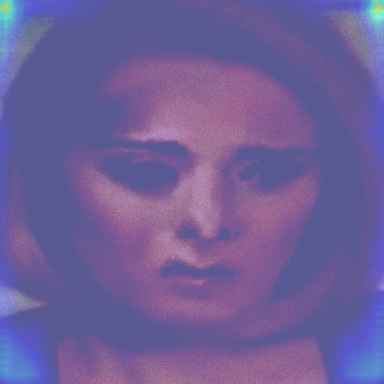} & 
        \includegraphics[scale=\pmmExampleScale]{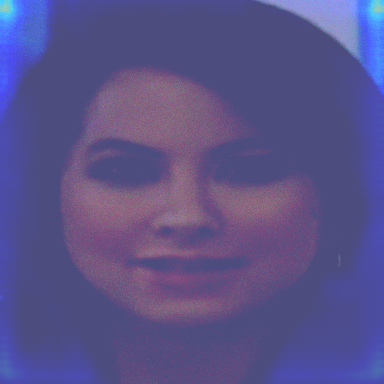} & 
        \includegraphics[scale=\pmmExampleScale]{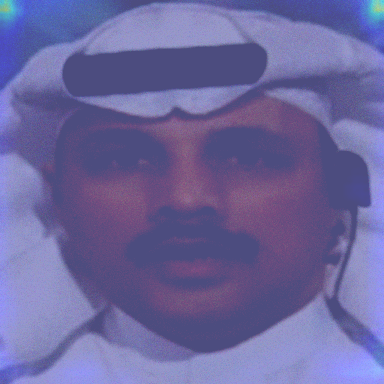} \\ \hline
        LAA+PMM (ours) & & 
        \includegraphics[scale=\pmmExampleScale]{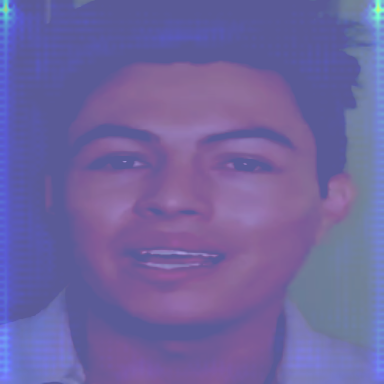} & 
        \includegraphics[scale=\pmmExampleScale]{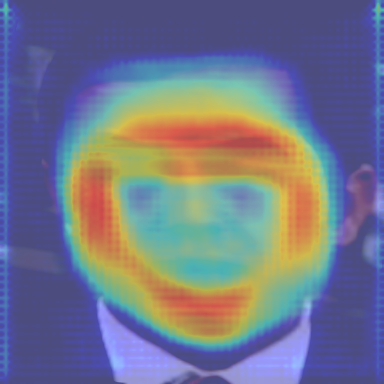} & 
        \includegraphics[scale=\pmmExampleScale]{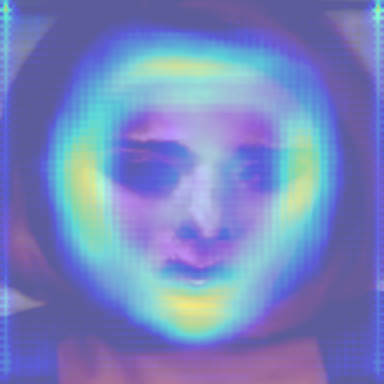} & 
        \includegraphics[scale=\pmmExampleScale]{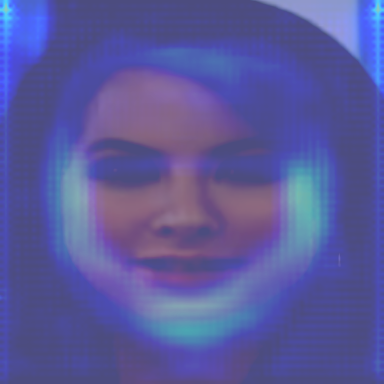} & 
        \includegraphics[scale=\pmmExampleScale]{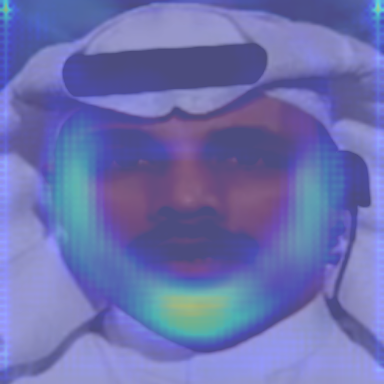} \\
        LAA+PMM (ours) & \checkmark & 
        \includegraphics[scale=\pmmExampleScale]{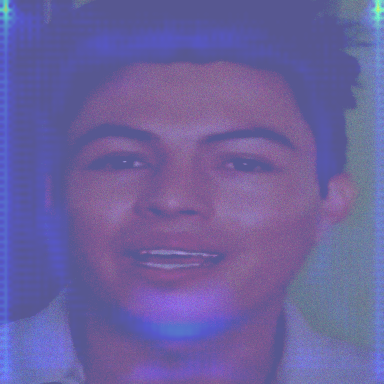} & 
        \includegraphics[scale=\pmmExampleScale]{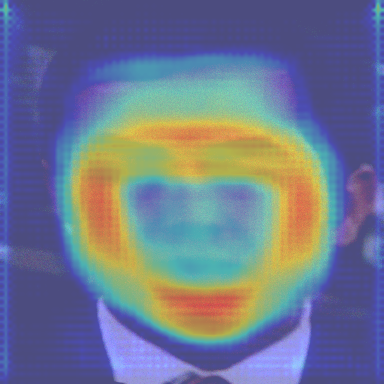} & 
        \includegraphics[scale=\pmmExampleScale]{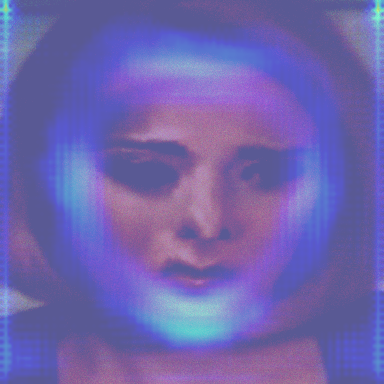} & 
        \includegraphics[scale=\pmmExampleScale]{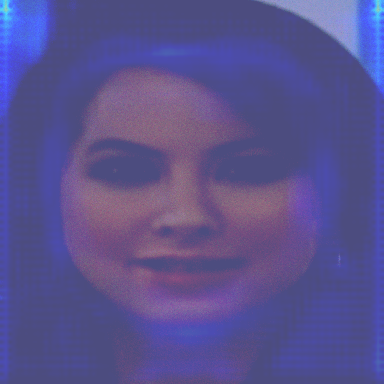} & 
        \includegraphics[scale=\pmmExampleScale]{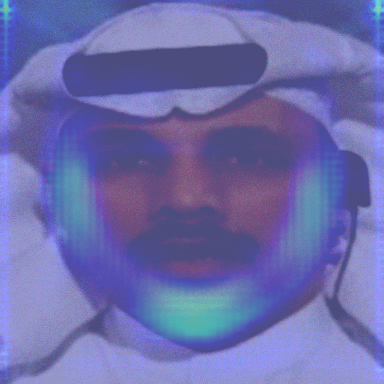} \\ \hline
    \end{tblr}
    \caption{\textbf{GradCAM visualization of our PMM compared to SBI~\protect\cite{Shiohara2022SBI} and LAA~\protect\cite{nguyen2024laa}.} Note that for noisy images, both SBI and LAA fail to capture useful information and therefore classify all images as \textcolor{green}{real}. Images are taken from the FaceForensics++ dataset~\protect\cite{roessler2019faceforensicspp}.}
    \label{tab:gradcam}
\end{table*}

\section{Additional robustness evaluation}
\label{sec:additional_robustness}

\begin{figure*}
	\centering
	\includegraphics[width=\linewidth]{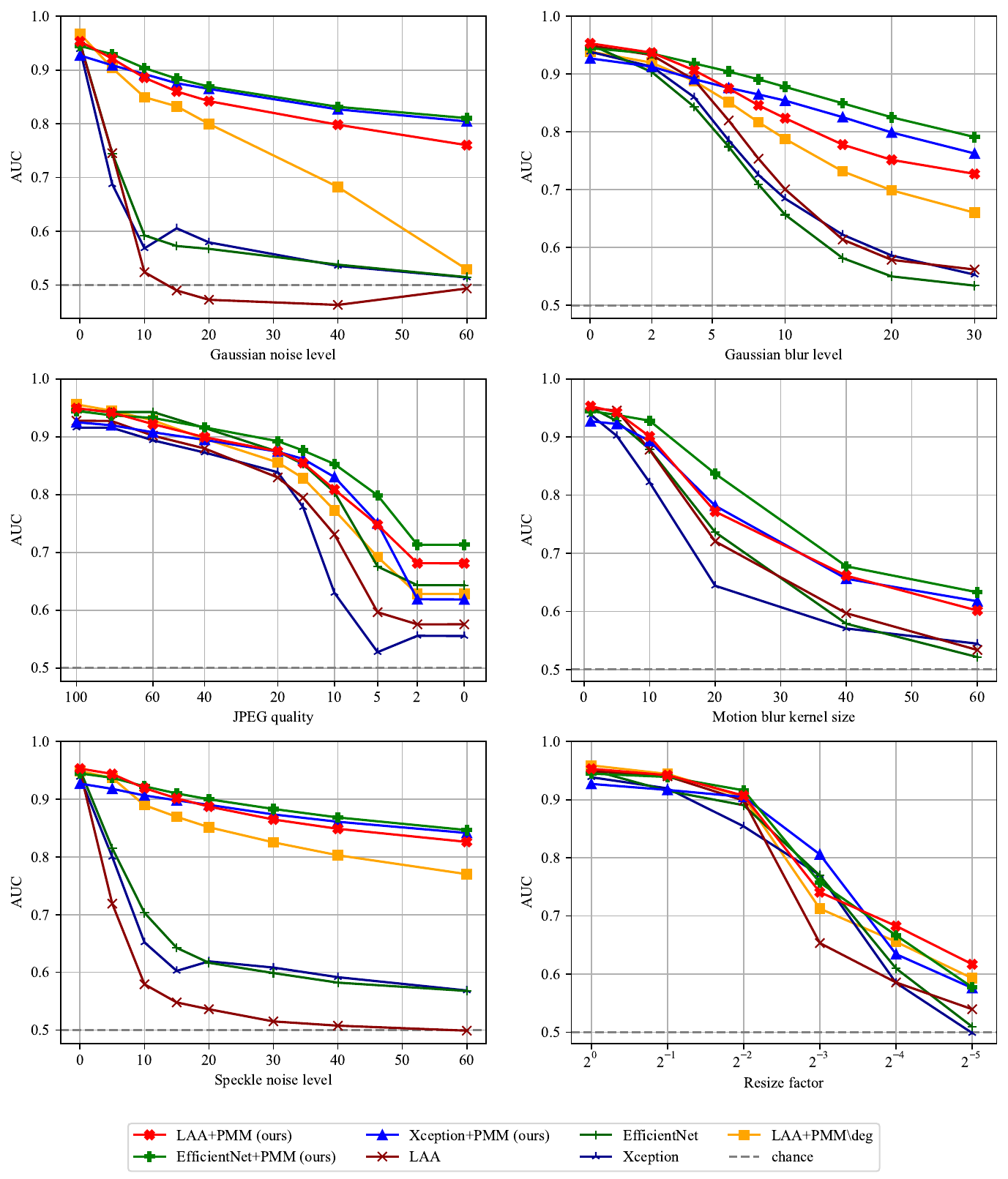}
	\caption{\textbf{Additional robustness evaluation} of the s-o-t-a deepfake detectors LAA~\protect\cite{nguyen2024laa} and SBI~\protect\cite{Shiohara2022SBI} (EfficientNet~\protect\cite{Tan2019EfficientNetRM} and Xception~\protect\cite{Chollet2016XceptionDL} backbones) vs. ours (+PMM), tested on the test-split of FF++~\protect\cite{roessler2019faceforensicspp}. The first four plots show the same data as \cref{fig:robustness} in the main paper. Similar to these, our models outperform the baselines for the low-quality settings of Speckle noise and Resizing.}
	\label{fig:robustness_supplementary}
\end{figure*}

We present an extended version of \cref{fig:robustness} in \cref{fig:robustness_supplementary}, where we additionally tested the robustness of LAA \cite{nguyen2024laa}, SBI \cite{Shiohara2022SBI} (EfficientNet \cite{Tan2019EfficientNetRM} and Xception \cite{Chollet2016XceptionDL} backbones), and ours (+PMM) to speckle noise and low resolution. For a description of the Gaussian noise, Gaussian blur, JPEG compression, and motion blur, see \cref{sec:robustness}. The diagram shown here is just an enlarged version for better readability.

\vspace{2mm}\noindent\textbf{Speckle noise.} Similar to the results for Gaussian noise, both SBI and LAA lose performance, even at small noise levels. At $\sigma = 20$, LAA performs at chance level ($0.54$ AUC), whereas SBI (both) is better at $>0.6$ AUC. However, our methods can beat both baselines by a large margin: SBI by $\geq 26\%$ and LAA by $\geq 35\%$. Even at the strongest settings $\sigma = 60$ (much larger than the $25 \cdot s = 12.5$ used during training), our methods still beat all baselines by $\geq 25\%$.

\vspace{2mm}\noindent\textbf{Resolution.} For large scale-factors of $\frac{1}{4}$ and above, all tested models are relatively robust, showing more than $0.89$ AUC, except for Xception, which only scores $0.85$ AUC. For small scale factors of $\frac{1}{16}$ and $\frac{1}{32}$, our model clearly outperforms the baselines, \eg by $7.7\%$, $10.8\%$ and $11.8\%$ for LAA+PMM vs. LAA, EfficientNet and Xception respectively at $\frac{1}{32}$.

\vspace{2mm}\noindent\textbf{Leave-one-out training.}
As described in the main paper, we also test models that use the entire PMM model, except for the specific test degradation, during training (PMM$\setminus$deg). They perform worse than the full PMM model, but better than LAA, indicating generalization to unseen degradations. 
Gaussian noise is similar but about three to 5 percent worse than PMM until $\sigma = 20$. For stronger noise, the no-Gaussian-noise model drops off significantly, reaching about chance level at $\sigma = 60$. This is still much better than LAA, which drops to chance performance by $\sigma = 10$.
For Gaussian blur, performance is better than LAA (up to $12\%$), but worse than full PMM (up to $6.7\%$).
JPEG compression also leads to a smaller performance drop than LAA, but larger than PMM (e.g. $68\%$ vs. $63\%$ vs. $58\%$ for PMM, PMM$\setminus$deg and LAA, respectively, at quality 0).
For speckle noise, performance is worse than PMM, by at most $5.6\%$, but much better than vanilla LAA, which is at chance performance, already at $\sigma \leq 10$. 
For small resize factors ($\leq \frac{1}{8}$), the model trained without resize is about $2.5\%$ worse than PMM. However, this is still about $6\%$ better than LAA.

These results suggest that due to the variety of degradations in our PMM, it is capable of generalizing to unseen degradations that were not part of the training data. This also aligns with our findings for motion blur. We therefore expect our model to generalize to novel degradations that we did not model during training.

\vspace{2mm}\noindent\textbf{New dataset.}
As describes in the main paper, our method can outperform LAA on the CDF based part of the face-swap and face-reenactment subsets of DF40~\cite{yan2024df40}. 
We also tested on the FF-based data. Here, LAA performs better than LAA+PMM (0.942 vs. 0.925 AUC). However, using data from the same source as the training data is not a true cross-dataset test and therefore not a good approximation to a real-world scenario, as it cannot be assumed that the real world follows the training distribution. The result is therefore only given for completeness.

\section{GradCAM evaluation}
\label{sec:gradcam}

Following related work \cite{nguyen2024laa, Shiohara2022SBI, Liu2021SPSL, Cao2022RECCE, Yan2023LSDA}, we visualize the inner workings of our model using GradCAM \cite{Selvaraju2016GradCAMVE}, comparing to SBI \cite{Shiohara2022SBI} and LAA \cite{nguyen2024laa}. The results are shown in \cref{tab:gradcam}. All models work well on clean images, but ours focuses more cleanly on the blending boundary, with very few other points in the image activated. This indicates that our method focuses almost exclusively on the artifact regions and neither on the face (like SBI) nor the background (like LAA). 

We also test on images, with added Gaussian noise ($\sigma = 20$), to explore the failure cases of the other models. In this case, both models focus on the image corners instead of the face. The focus of our model barely changes and only loses some strength. This is consistent with the observation that we can still correctly classify the fake images, whereas LAA and SBI cannot.

\section{Qualitative demonstration}
\label{sec:qualitative}

\begin{figure}
    \centering
    \begin{subfigure}{\linewidth}  
        \centering
        \includegraphics[width=0.5\linewidth]{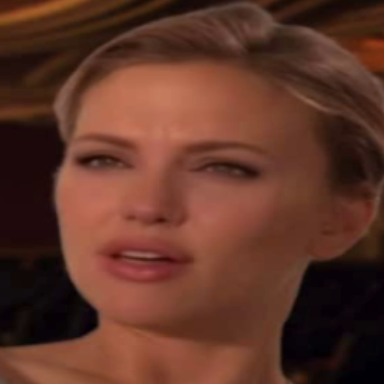}
        \caption{\textcolor{red}{Fake} image: LAA~\protect\cite{nguyen2024laa} predicts \textcolor{green}{real}, ours predicts \textcolor{red}{fake}.}
        \label{fig:laa_failure}
    \end{subfigure}
    
    \begin{subfigure}{\linewidth}  
        \centering
        \includegraphics[width=0.5\linewidth]{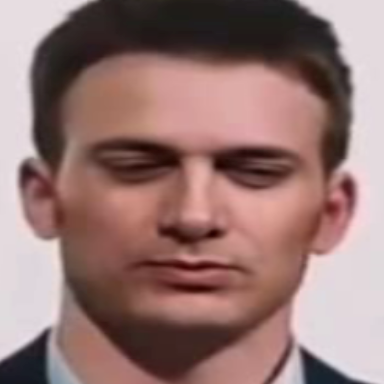}
        \caption{\textcolor{red}{Fake} image: LAA~\protect\cite{nguyen2024laa} predicts \textcolor{green}{real}, ours predicts \textcolor{green}{real}.}
        \label{fig:both_failure}
    \end{subfigure}
    
    \begin{subfigure}{\linewidth}  
        \centering
        \includegraphics[width=0.5\linewidth]{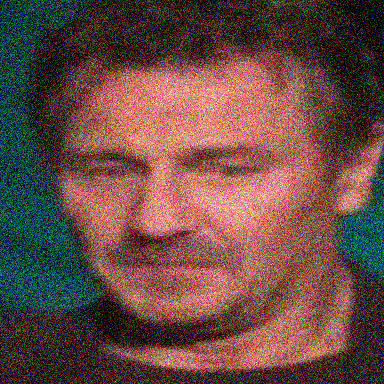}
        \caption{\textcolor{red}{Fake} image: LAA~\protect\cite{nguyen2024laa} predicts \textcolor{green}{real}, ours predicts \textcolor{red}{fake}.}
        \label{fig:hard}
    \end{subfigure}
    \caption{\textbf{Qualitative demonstration of LAA~\protect\cite{nguyen2024laa} and our model.} All images are taken from the Celeb-DF-v2 dataset~\protect\cite{Li2019CelebDFv2}. For image (c), we added strong Gaussian noise to showcase the robustness of our model.}
    \label{fig:qualitative}
\end{figure}

\Cref{fig:qualitative} shows additional qualitative evaluations of LAA \cite{nguyen2024laa} and our model. All images are taken from the Celeb-DF-v2 dataset \cite{Li2019CelebDFv2}.

\Cref{fig:laa_failure} shows a failure case for LAA. Despite the image being sourced from the high-quality Celeb-DF-v2 dataset, the image is slightly blurry. LAA fails to recognize the image as fake, whereas ours can recognize it.

\Cref{fig:both_failure} shows a failure case for both models. Again, the image is slightly blurry, but it also seems to be a very good fake, as neither LAA nor our model can recognize it as fake. From a human perspective, we also cannot tell any visible signs of the image being fake. This emphasizes the need to also continue developing models towards high-quality fakes. In this paper, we mainly concern ourselves with robustness to low-quality data.

Finally, \cref{fig:hard} shows an example of the increased robustness of our model. We add Gaussian noise ($\sigma = 40$), and yet, our model can still recognize the image as fake, while LAA considers it real. Again, from a human perspective, we cannot tell a sign of a fake, given that the image is extremely noisy.

\section{Recognizability}
\label{sec:recognizability}

\begin{figure}[h!]
	\centering
	\includegraphics[width=0.6\linewidth]{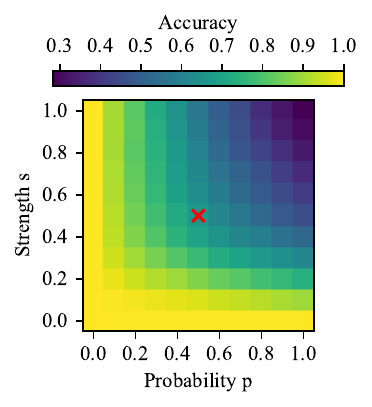}
	\caption{Recognizability of faces under our degradation model. \color{red}\xmark\color{black}\ marks our settings during PMM training.}
	\label{fig:facedetect_PMM}
\end{figure}

\begin{figure}
	\centering
	\includegraphics[width=\linewidth]{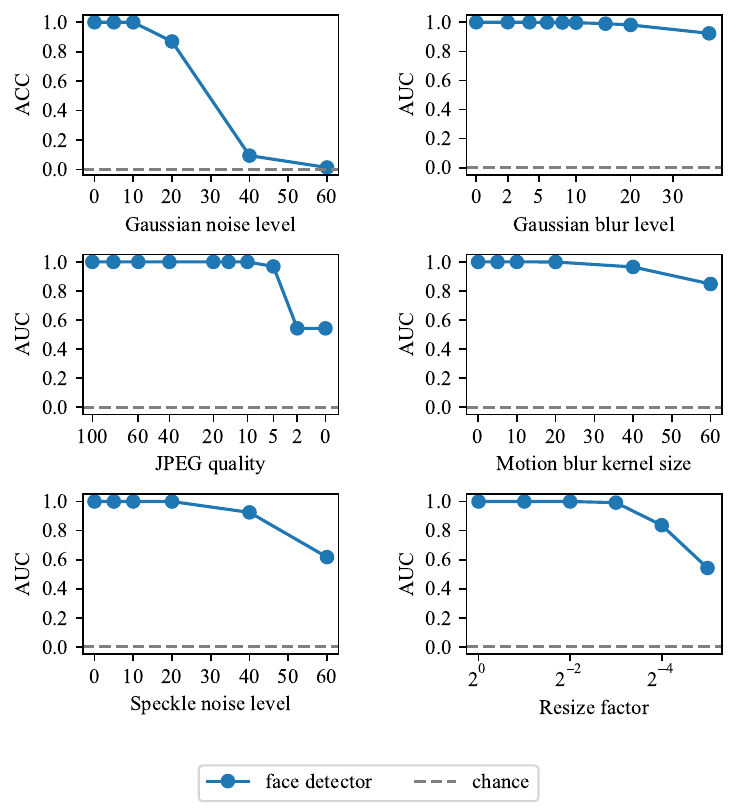}
	\caption{Robustness of the face detector to the degradations from \cref{fig:robustness} and \cref{fig:robustness_supplementary}}
	\label{fig:facedetect_individual}
\end{figure}

To measure the difficulty of our degradations, we provide a metric that shows how recognizable faces remain. For this purpose, we train an EfficientNet-b4 \cite{Tan2019EfficientNetRM} to recognize the identities of the faces in the FaceForensics++ \cite{roessler2019faceforensicspp} train split (720 identities). Then, we test the model under several kinds of degradations. \Cref{fig:facedetect_PMM} shows the results under our degradation model. At the PMM settings, the face detector still achieves an accuracy of $67\%$, so our settings are of reasonable hardness. Furthermore, we can see (\cref{fig:facedetect_individual}) that most individual degradations shown in Figures \ref{fig:robustness} and \ref{fig:robustness_supplementary} still allow for an accuracy of $>50\%$ (at 720 possible identities). The only exception is noise; however, here, the face detector is still robust (accuracy $> 85\%$) until a value of $\sigma = 20$, where previous deepfake detectors already fail. 

We can, therefore, show that our degradations are of reasonable difficulty, as faces still remain recognizable under these conditions.

\section{Implementation details}
\label{sec:implementation}
We mainly use the settings from LAA~\cite{nguyen2024laa} with some exceptions. 
We use a batch size of 7 to fit into the 24GB of VRAM of the NVIDIA RTX 3090 that we use for training.
Furthermore, we use a learning rate scheduler, which reduces the learning rate by a factor of $0.2$ every time the validation loss does not decrease for 10 epochs. We make this change to avoid having to tune the schedule for our experiments manually. 
Since we do not make any changes to the model itself, we can start from the pretrained checkpoint provided by~\cite{nguyen2024laa}, to save compute and ensure we start from the s-o-t-a point.
	
\end{document}